\documentclass[sigconf]{acmart}

\usepackage{algorithm}
\usepackage{algpseudocode}
\usepackage{amsmath}
\usepackage{array}
\usepackage{booktabs}
\usepackage{enumitem}
\usepackage{graphicx}
\usepackage{listings}
\usepackage{multirow}
\usepackage{pifont}
\usepackage{subcaption}
\usepackage{xcolor}
\usepackage{xspace}

\newcommand{\dataspace}{\textsc{DataSpace}\xspace}
\newcommand{\dataspacebuilder}{\textsc{DataSpace-Builder}\xspace}
\newcommand{\dataspaceagent}{\textsc{DataSpace-Agent}\xspace}

\newcommand{\workspace}{\mathcal{W}}
\newcommand{\goldrel}{\mathcal{Y}}
\newcommand{\prediction}{\widehat{\mathcal{Y}}}
\newcommand{\database}{\mathcal{D}}

\definecolor{tablecheck}{HTML}{216D68}
\newcommand{\yesmark}{\textcolor{tablecheck}{\ding{51}}}
\newcommand{\nomark}{\textcolor{gray}{--}}
\newcommand{\tablehead}[1]{%
  {\bfseries\begin{tabular}[c]{@{}c@{}}#1\end{tabular}}%
}

\definecolor{docanchor}{HTML}{B45309}
\definecolor{doccontext}{HTML}{1D4ED8}
\definecolor{docmeasure}{HTML}{047857}

\definecolor{findingaccent}{HTML}{216D68}
\definecolor{findingbackground}{HTML}{EFF7F5}
\newcommand{\findingbox}[2]{%
  \par\smallskip
  \noindent
  \begingroup
  \setlength{\fboxsep}{5pt}%
  \colorbox{findingbackground}{%
    \parbox{\dimexpr\columnwidth-2\fboxsep\relax}{%
      \small\textcolor{findingaccent}{\textbf{Finding #1.}}\enspace #2%
    }%
  }%
  \endgroup
  \par\smallskip
}

\newenvironment{promptblock}[2]
  {\begin{figure}[t]\centering
   \def\promptcaption{#1}\def\promptlabel{#2}%
   \begin{minipage}{\columnwidth}}
  {\end{minipage}
   \caption{\promptcaption}\label{\promptlabel}
   \end{figure}}

\graphicspath{{figures/}}

\microtypesetup{protrusion=false}
\setcopyright{none}
\renewcommand\footnotetextcopyrightpermission[1]{}
\begin{document}

\title[DataSpace]{DataSpace: Benchmarking Data Agents for Verifiable Analytics
over Heterogeneous Workspaces}

\author{Boyan Li}
\authornote{These authors contributed equally to this work.}
\affiliation{%
  \institution{HKUST(GZ)}
  \city{Guangzhou}
  \country{China}}

\author{Zhuowen Liang}
\authornotemark[1]
\affiliation{%
  \institution{HKUST(GZ)}
  \city{Guangzhou}
  \country{China}}

\author{Yupeng Xie}
\authornotemark[1]
\affiliation{%
  \institution{HKUST(GZ)}
  \city{Guangzhou}
  \country{China}}

\author{Xiaotian Lin}
\affiliation{%
  \institution{HKUST(GZ)}
  \city{Guangzhou}
  \country{China}}

\author{Tianqi Luo}
\affiliation{%
  \institution{HKUST(GZ)}
  \city{Guangzhou}
  \country{China}}

\author{Xinyu Liu}
\affiliation{%
  \institution{HKUST(GZ)}
  \city{Guangzhou}
  \country{China}}

\author{Yizhang Zhu}
\affiliation{%
  \institution{HKUST(GZ)}
  \city{Guangzhou}
  \country{China}}

\author{Zhangyang Peng}
\affiliation{%
  \institution{HKUST(GZ)}
  \city{Guangzhou}
  \country{China}}

\author{Yuan Li}
\affiliation{%
  \institution{Tsinghua University}
  \city{Beijing}
  \country{China}}

\author{Zhengxuan Zhang}
\affiliation{%
  \institution{HKUST(GZ)}
  \city{Guangzhou}
  \country{China}}

\author{Jiayi Zhang}
\affiliation{%
  \institution{HKUST(GZ)}
  \city{Guangzhou}
  \country{China}}

\author{Nan Tang}
\affiliation{%
  \institution{HKUST(GZ)}
  \city{Guangzhou}
  \country{China}}

\author{Guoliang Li}
\affiliation{%
  \institution{Tsinghua University}
  \city{Beijing}
  \country{China}}

\author{Yuyu Luo}
\affiliation{%
  \institution{HKUST(GZ)}
  \city{Guangzhou}
  \country{China}}

\renewcommand{\shortauthors}{Li et al.}

\begin{abstract}
Data agents enable natural-language analytics over organizational workspaces,
where relevant evidence may be scattered across databases, structured files,
long documents, and multimedia. Existing benchmarks largely isolate structured
querying, retrieval, or open-ended analysis, leaving heterogeneous evidence
discovery, complete tabular outputs, and deterministic evaluation
insufficiently unified. We introduce \dataspace, a benchmark in which data
agents produce verifiable tabular results from task-local heterogeneous
workspaces. It contains 410
cross-language tasks and 7,439 artifacts totaling 15.01~GB across CSV, JSON,
SQLite, Markdown, PDF, and video.
\dataspace also served as the official evaluation benchmark for the KDD Cup
2026 Data Agents for Complex Data Analysis competition. Each agent receives
only a question and workspace and returns the complete requested tabular
result. We construct \dataspace with \dataspacebuilder, an
execution-grounded framework comprising cross-language transformation,
constraint-aware relational sampling, modality routing and artifact
rendering, and human review and task repair by 11 domain experts. A
deterministic evaluator performs
header-invariant column alignment, type- and precision-aware normalization,
and order-aware row comparison. Across six recently released frontier
multimodal models and five widely used agent harnesses, the best accuracy
reaches 66.34\%, while harness choice creates a 15.36-point spread with the
backbone fixed.
Multimodal evidence integration and joins consistently reduce accuracy across
all six backbones. These results show that \dataspace remains unsaturated
and identify key challenges for improving data-agent reliability.
\end{abstract}


\maketitle

\section{Introduction}
\label{sec:introduction}

\begin{figure}[!t]
    \centering
    \includegraphics[width=\columnwidth]{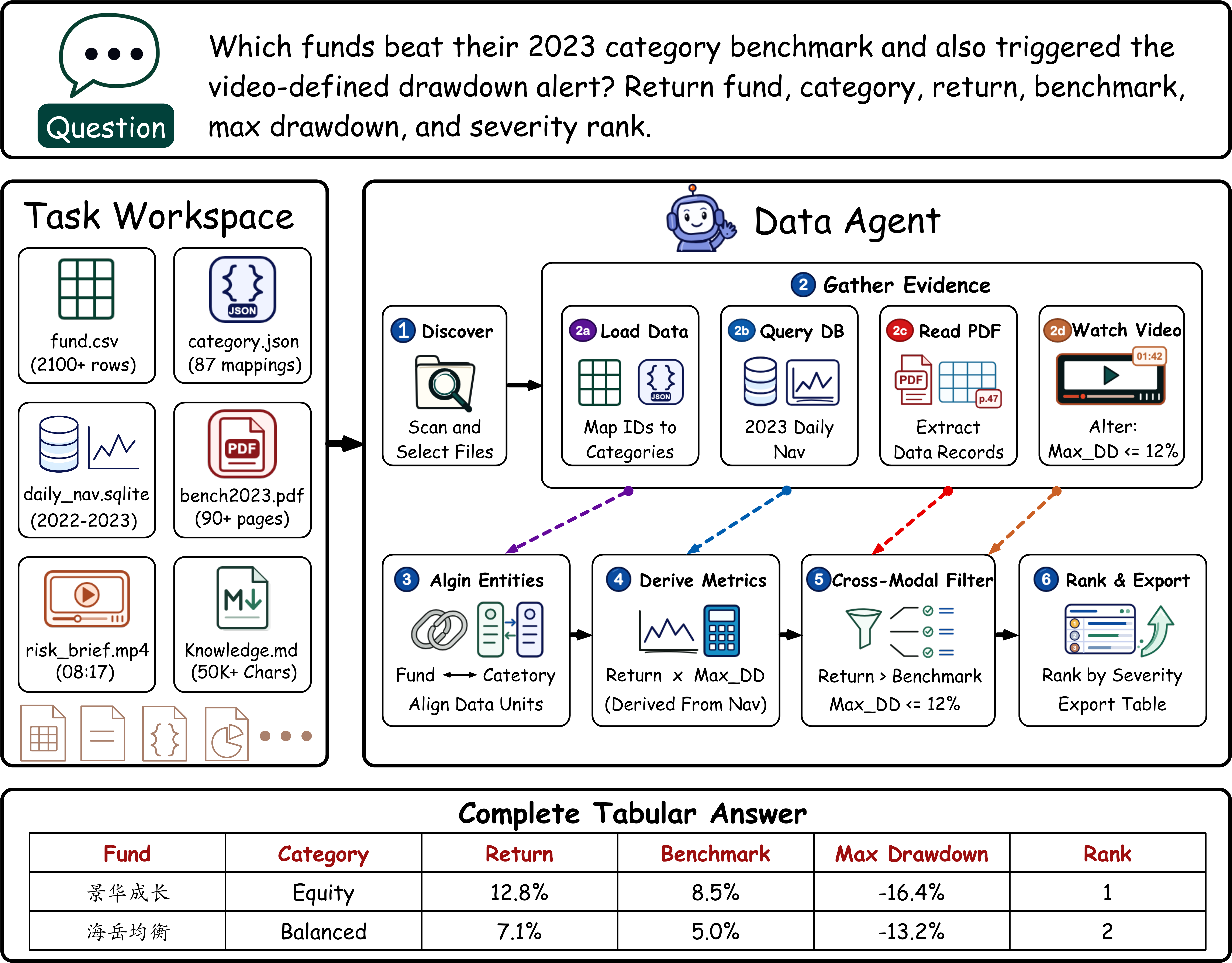}
    \caption{The \dataspace task interface, illustrated with a fund-risk task.
    The agent combines an alert rule from video, category benchmarks extracted
    from a long PDF, and daily NAV queried from SQLite, then aligns entities,
    computes the requested metrics, and returns the complete tabular result.}
    \label{fig:task-example}
\end{figure}

Data agents are emerging as a natural-language interface to organizational
data~\cite{aop, agenticdata, deepanalyze, taiji, teable, deepeye,
dataagentsurvey,deepvis,nvbench2,xie2024haichart,tang2026vividoc}. In realistic analytical settings, however, the information needed to
answer a user question rarely resides in a single clean table or a preselected
database~\cite{fdabench, datacross, dabstep}.
%
\begin{table*}[!t]
    \caption{Comparison of representative benchmarks in artifact coverage,
    workspace demands, and answer/evaluation semantics.}
    \label{tab:benchmark-comparison}
    \centering
    \scriptsize
    \setlength{\tabcolsep}{2.0pt}
    \renewcommand{\arraystretch}{0.92}
    \setlength{\aboverulesep}{0.25ex}
    \setlength{\belowrulesep}{0.40ex}
    \setlength{\cmidrulesep}{0.50ex}
    \resizebox{\textwidth}{!}{%
    \begin{tabular}{@{}l l r c c c c c c c c c c c c@{}}
        \toprule
        \multirow[c]{2}{*}[-1.0ex]{\tablehead{Type}}
        & \multirow[c]{2}{*}[-1.0ex]{\tablehead{Benchmark}}
        & \multirow[c]{2}{*}[-1.0ex]{\tablehead{\#Tasks}}
        & \multicolumn{4}{c}{\textbf{Input artifacts}}
        & \multicolumn{5}{c}{\textbf{Workspace requirements}}
        & \multicolumn{3}{c}{\textbf{Output and evaluation}} \\
        \cmidrule(lr){4-7}\cmidrule(lr){8-12}\cmidrule(l){13-15}
        & & & \tablehead{DB} & \tablehead{Files} & \tablehead{Docs}
        & \tablehead{Media} & \tablehead{Cross-art.}
        & \tablehead{Discovery} & \tablehead{Long\\docs}
        & \tablehead{Doc$\rightarrow$\\records} & \tablehead{Cross-lang.}
        & \tablehead{Complete\\table} & \tablehead{Model-\\free}
        & \tablehead{Schema-\\inv.} \\
        \midrule
        \multirow{4}{*}{Structured}
        & WikiTableQuestions~\cite{wikitablequestions}
        & 22,033
        & \nomark & \yesmark & \nomark & \nomark
        & \nomark & \nomark & \nomark & \nomark & \nomark
        & \nomark & \yesmark & \nomark \\
        & Spider~\cite{spider}
        & 10,181
        & \yesmark & \nomark & \nomark & \nomark
        & \nomark & \nomark & \nomark & \nomark & \nomark
        & \yesmark & \yesmark & \nomark \\
        & BIRD~\cite{bird}
        & 12,751
        & \yesmark & \nomark & \nomark & \nomark
        & \nomark & \nomark & \nomark & \nomark & \nomark
        & \yesmark & \yesmark & \nomark \\
        & Spider~2.0~\cite{spider2}
        & 632
        & \yesmark & \nomark & \yesmark & \nomark
        & \yesmark & \yesmark & \nomark & \nomark & \nomark
        & \yesmark & \yesmark & \nomark \\
        \midrule
        \multirow{3}{*}{Unstructured}
        & HotpotQA~\cite{hotpotqa}
        & 113K
        & \nomark & \nomark & \yesmark & \nomark
        & \yesmark & \yesmark & \nomark & \nomark & \nomark
        & \nomark & \yesmark & \nomark \\
        & CRAG~\cite{crag}
        & 4,409
        & \nomark & \nomark & \yesmark & \nomark
        & \yesmark & \yesmark & \nomark & \nomark & \nomark
        & \nomark & \nomark & \nomark \\
        & MMLongBench-Doc~\cite{mmlongbenchdoc}
        & 1,062
        & \nomark & \nomark & \yesmark & \nomark
        & \nomark & \nomark & \yesmark & \nomark & \nomark
        & \nomark & \yesmark & \nomark \\
        \midrule
        \multirow{5}{*}{Data agent}
        & DABStep~\cite{dabstep}
        & 450+
        & \nomark & \yesmark & \yesmark & \nomark
        & \yesmark & \yesmark & \nomark & \nomark & \nomark
        & \nomark & \yesmark & \nomark \\
        & KramaBench~\cite{kramabench}
        & 104
        & \nomark & \yesmark & \yesmark & \nomark
        & \yesmark & \yesmark & \nomark & \nomark & \nomark
        & \nomark & \nomark & \nomark \\
        & LongDA~\cite{longda}
        & 505
        & \nomark & \yesmark & \yesmark & \nomark
        & \yesmark & \yesmark & \yesmark & \nomark & \nomark
        & \nomark & \yesmark & \nomark \\
        & DataCross~\cite{datacross}
        & 200
        & \yesmark & \yesmark & \yesmark & \nomark
        & \yesmark & \yesmark & \nomark & \yesmark & \nomark
        & \nomark & \nomark & \nomark \\
        & FDABench~\cite{fdabench}
        & 2,007
        & \yesmark & \yesmark & \yesmark & \yesmark
        & \yesmark & \yesmark & \nomark & \nomark & \nomark
        & \nomark & \nomark & \nomark \\
        \midrule
        \textbf{Ours}
        & \textbf{\dataspace}
        & \textbf{410}
        & \yesmark & \yesmark & \yesmark & \yesmark
        & \yesmark & \yesmark & \yesmark & \yesmark & \yesmark
        & \yesmark & \yesmark & \yesmark \\
        \bottomrule
    \end{tabular}%
    }

    \vspace{0pt}
    \parbox{\textwidth}{\scriptsize
        \textit{Notes.} Input/workspace checkmarks denote explicit coverage;
        output/evaluation checkmarks denote benchmark-wide requirements.
        Files: standalone structured/semi-structured artifacts; Docs:
        textual/visual documents;
        Media: audio/video. Cross-art.: multi-artifact/system integration;
        Discovery: sources not preselected; Long docs: explicit long-document
        processing; Doc$\rightarrow$records: document fields/records feed
        downstream analysis; Cross-lang.: joint question--workspace input.
        Complete table: correctness requires the complete result table,
        submitted directly or obtained by query execution; Model-free: no LLM
        judge; Schema-inv.:
        columns align despite header wording or order.}
\end{table*}

The question and its evidence may cross languages and representations,
spanning relational databases, structured and semi-structured files, business
documents, and multimedia artifacts alongside valid but irrelevant files. An
effective data agent therefore acts as a \emph{workspace solver}: it inspects
the available data, selects sources and tools, aligns information across
representations, executes multi-step computations, and returns a result that
the user can directly consume.

Existing benchmarks capture complementary parts of this setting.
\emph{Structured-data benchmarks}, from Spider~\cite{spider} to
Spider~2.0~\cite{spider2}, offer strong tests of relational reasoning with
deterministic evaluation, but generally identify the relevant table or
database in advance. \emph{Unstructured-data benchmarks} such as
MMLongBench-Doc~\cite{mmlongbenchdoc} introduce long, visually rich inputs,
yet focus primarily on evidence retrieval, grounding, and answer synthesis.
\emph{Data-agent benchmarks} move closer to open-ended analytical workspaces:
KramaBench~\cite{kramabench} studies data-to-insight pipelines over data lakes,
while FDABench~\cite{fdabench} extends analysis across structured data,
documents, and media. As summarized in
Table~\ref{tab:benchmark-comparison}, these advances have not yet unified
three properties central to realistic data analysis:
\textbf{(L1)~Workspace scope:} a task-local workspace spanning structured
files, databases, long documents, and multimedia artifacts, with language
variation across both the question and data;
\textbf{(L2)~Output contract:} a consistent objective requiring the complete
analytical result rather than a factoid, pipeline, or open-ended report; and
\textbf{(L3)~Evaluation semantics:} deterministic evaluation that accepts
equivalent representations while rejecting incomplete or erroneous answers.

We therefore introduce \dataspace\footnote{%
\begin{tabular}[t]{@{}l@{\enspace}p{0.74\linewidth}@{}}
Code: & \url{https://github.com/HKUSTDial/DataSpace}\\
Dataset: & \url{https://huggingface.co/datasets/HKUSTDial/DataSpace}
\end{tabular}},
a benchmark for verifiable data analytics
over self-contained heterogeneous workspaces
(Figure~\ref{fig:task-example}). An agent receives only a natural-language
question and a task-local workspace, autonomously discovers and combines the
available data, and returns the complete requested tabular result. \dataspace
contains 410 cross-language tasks and 7,439 artifacts totaling approximately
15\,GB.
Chinese and English may occur across both the user question and its workspace
artifacts. The workspaces span CSV, JSON, SQLite, Markdown, PDF, and Video,
and each task is paired with a complete tabular reference answer. \textbf{\dataspace
also served as the official evaluation benchmark for the KDD Cup 2026 Data
Agents for Complex Data Analysis competition}~\cite{kddcup2026dataagents}.

To construct such cross-language, heterogeneous workspaces reliably, we
propose \dataspacebuilder, an execution-grounded framework that transforms
instances from EHRSQL~\cite{ehrsql} and BULL~\cite{finsql}, two English
Text-to-SQL benchmarks covering clinical and financial analytics. Their
relational databases supply domain data, while executable SQL provides
parseable analytical logic and execution-based validation. \dataspacebuilder
comprises four stages:
\emph{Cross-Language Transformation}, \emph{Constraint-Aware Relational
Sampling}, \emph{Modality Routing \& Artifact Rendering}, and
\emph{Human Review \& Task Repair}. Together, these stages
transform source questions, databases, and query logic into cross-language,
heterogeneous workspace tasks and derive their reference answers through
execution. The resulting question, workspace, reference answer, and
evaluation semantics are cross-reviewed by a panel of 11 domain experts, with
disputed cases discussed and repaired before release. We also design a
deterministic evaluator that aligns columns regardless of header wording or
position, normalizes equivalent value formats, and compares rows according to
whether their order matters for the task.

\noindent\textbf{Contributions.}
Our contributions are:
\begin{itemize}[leftmargin=*]
    \item \textbf{A heterogeneous workspace benchmark.} We introduce
    \dataspace, comprising 410 cross-language tasks and six modalities
    under a uniform, exactly verifiable tabular-output objective.
    \item \textbf{An execution-grounded construction framework.} We develop
    \dataspacebuilder, which transforms executable Text-to-SQL resources into
    task-local heterogeneous workspaces with expert review and repair.
    \item \textbf{A semantics-aware evaluator.} We provide deterministic
    evaluation of complete tabular results that tolerates equivalent
    representations while rejecting incomplete or erroneous outputs.
    \item \textbf{An empirical study of data agents.} We establish
    baselines across six frontier multimodal models and five agent harnesses;
    the best reaches 66.34\% accuracy, harness choice produces a 15.36-point
    spread, and multimodal evidence and joins consistently reduce performance.
\end{itemize}

\section{Related Work}
\label{sec:related-work}

\noindent\textbf{Structured-data benchmarks.}
Natural-language analytics over structured data is studied through
table question answering and Text-to-SQL~\cite{dawnnl2sql,alphasql,
deepeyesql,dpc,rose,sqlconductor,nl2sqlrewriter,texttosqllmsurvey,
nl2sqlbugs,elliesql}. Table QA benchmarks such as
WikiTableQuestions~\cite{wikitablequestions} predict denotations over
semi-structured tables. Spider~\cite{spider} and BIRD~\cite{bird} cover
cross-domain and large-scale databases, while EHRSQL~\cite{ehrsql} and
BULL~\cite{finsql} capture clinical and financial analytics.
Spider~2.0~\cite{spider2} further introduces enterprise artifacts around the
SQL workflow.
Executable queries enable deterministic result-level evaluation, but these
settings provide limited coverage of evidence discovery and reconciliation
across separate files, documents, and media.

\noindent\textbf{Unstructured-data benchmarks.}
HotpotQA~\cite{hotpotqa} and CRAG~\cite{crag} test retrieval and synthesis
over multi-hop or retrieval-augmented corpora, while
FinanceBench~\cite{financebench} and
MMLongBench-Doc~\cite{mmlongbenchdoc,liang2026long} target financial reports and long,
visually rich documents. HybridQA~\cite{hybridqa},
MultiModalQA~\cite{multimodalqa}, and Video-MME~\cite{videomme} add linked
tables, passages, images, or video. These settings provide strong tests of
evidence localization, cross-page reasoning, and perception, but generally
target factoids, short lists, choices, or free-form responses. They rarely
require recovering typed record collections from long documents and combining
them with other workspace data to produce a complete tabular result.

\noindent\textbf{Data-agent benchmarks.}
DABStep~\cite{dabstep}, KramaBench~\cite{kramabench}, and
LongDA~\cite{longda,bian2025you} cover multistep processing, data-lake
discovery, and long-document navigation. Data Agent
Benchmark~\cite{dataagentbenchmark} and
AgenticDataBench~\cite{agenticdatabench} emphasize multi-system querying or
recurring data-science skills, while DataCross~\cite{datacross} and
FDABench~\cite{fdabench,chen2025chartmark,zhang2025datamosaic} incorporate visual tables and media. This
family is closest to our setting, but its targets range from factoid answers
and executable pipelines to choices and reports, accompanied by execution-,
rubric-, or model-based evaluation~\cite{xie2025visjudge,tang2026igenbench}.
\dataspace instead holds the task contract
fixed: every task requires a complete tabular result scored by the same
deterministic protocol.

\section{Benchmark Overview and Task Formulation}
\label{sec:benchmark}

\dataspace evaluates a data agent as a workspace solver: given an analytical
question and the contents of a task-local workspace, the agent must return the
complete typed table requested by the user. This section defines the task
interface, summarizes the benchmark scope, and identifies the capabilities
exercised by this setting.

\subsection{Task Formulation}

For task \(i\), the public input is
\begin{equation}
    x_i=(q_i,\workspace_i), \qquad
    \workspace_i=\workspace_i^{\mathrm{str}}\cup
    \workspace_i^{\mathrm{doc}}\cup\workspace_i^{\mathrm{med}},
    \label{eq:task-interface}
\end{equation}
where \(q_i\) is a natural-language question. The workspace contains
structured and semi-structured artifacts
\(\workspace_i^{\mathrm{str}}\) (CSV, JSON, and SQLite), document artifacts
\(\workspace_i^{\mathrm{doc}}\) (Markdown and PDF), and media artifacts
\(\workspace_i^{\mathrm{med}}\) (video). The initial observation \(o_0\)
gives the agent the question and access to the workspace root; artifact
contents are acquired as the agent interacts with the workspace.

Let \(\mathcal{T}_A\) denote the tools available to agent \(A\). They may
include file-system inspection, structured-data parsing, SQL execution, code
execution, document extraction, and video understanding. The action space,
agent state, and nonterminal environment transition are jointly defined as
\begin{equation}
    \begin{aligned}
        \mathcal{A}_A
            &= \{\operatorname{Call}(\tau,\theta):
                 \tau\in\mathcal{T}_A,\theta\in\Theta_\tau\} \\
            &\quad\cup\{\operatorname{Answer}(Y):
                 Y\text{ is a tabular result}\}, \\
        h_t&=(o_0,a_0,\ldots,a_{t-1},o_t),\qquad s_t=(h_t,m_t), \\
        a_t&=\pi_A(s_t),\qquad
        (o_{t+1},m_{t+1})=\operatorname{Exec}_i(s_t,a_t),
    \end{aligned}
    \label{eq:agent-interaction}
\end{equation}
where \(\Theta_\tau\) is the argument space of tool \(\tau\), \(m_t\) is
working memory, and
\(\pi_A\) is the agent policy. For a tool-call action,
\(\operatorname{Exec}_i\) executes the selected tool in the fixed workspace
and returns its observation. The terminal action is
\(a_T=\operatorname{Answer}(\prediction_i)\), where \(\prediction_i\) is a
tabular result serialized as a CSV file.
Figure~\ref{fig:task-example} illustrates
the resulting task-level input--output interface.

The benchmark-side record additionally contains the reference result
\(\goldrel_i\) and a compact evaluation configuration \(c_i\). The benchmark
record and binary task score are
\begin{equation}
    b_i=(x_i,\goldrel_i,c_i), \qquad
    s_i=\mathbf{1}\!\left[\prediction_i\equiv_{c_i}\goldrel_i\right].
    \label{eq:task-correctness}
\end{equation}
The configuration records the semantic type of each reference column,
numeric comparison rules where needed, and whether row order is significant.
It belongs to the evaluation protocol rather than the question--workspace
input. A task is correct when its complete tabular prediction matches the
reference result under these semantics; missing or extra rows and columns
make the prediction incorrect even when some returned values match.
Section~\ref{sec:evaluation} specifies the equivalence criterion and the
aggregate metric.

\subsection{Benchmark Scope}

\dataspace contains 410 tasks spanning financial, macroeconomic, and
healthcare analytics. Its workspaces combine CSV, JSON, SQLite, Markdown, PDF,
and video in 13 modality combinations, and questions and artifacts may mix
Chinese and English. Every sampled source table remains represented: long
documents may encode records or fields, while video may supply a condition,
intermediate value, or answer. Answer tables range from a single cell to
multi-column outputs with thousands of rows.

Tasks couple four capabilities: \textbf{(i) workspace discovery};
\textbf{(ii) interpretation and alignment} of types, schemas, entities, units,
and languages; \textbf{(iii) relational computation}, such as filtering,
joining, aggregation, ranking, and temporal reasoning; and
\textbf{(iv) complete tabular-result materialization}.

\section{Benchmark Construction and Evaluation}
\label{sec:construction}

\begin{figure*}[t]
    \centering
    \includegraphics[width=\textwidth]{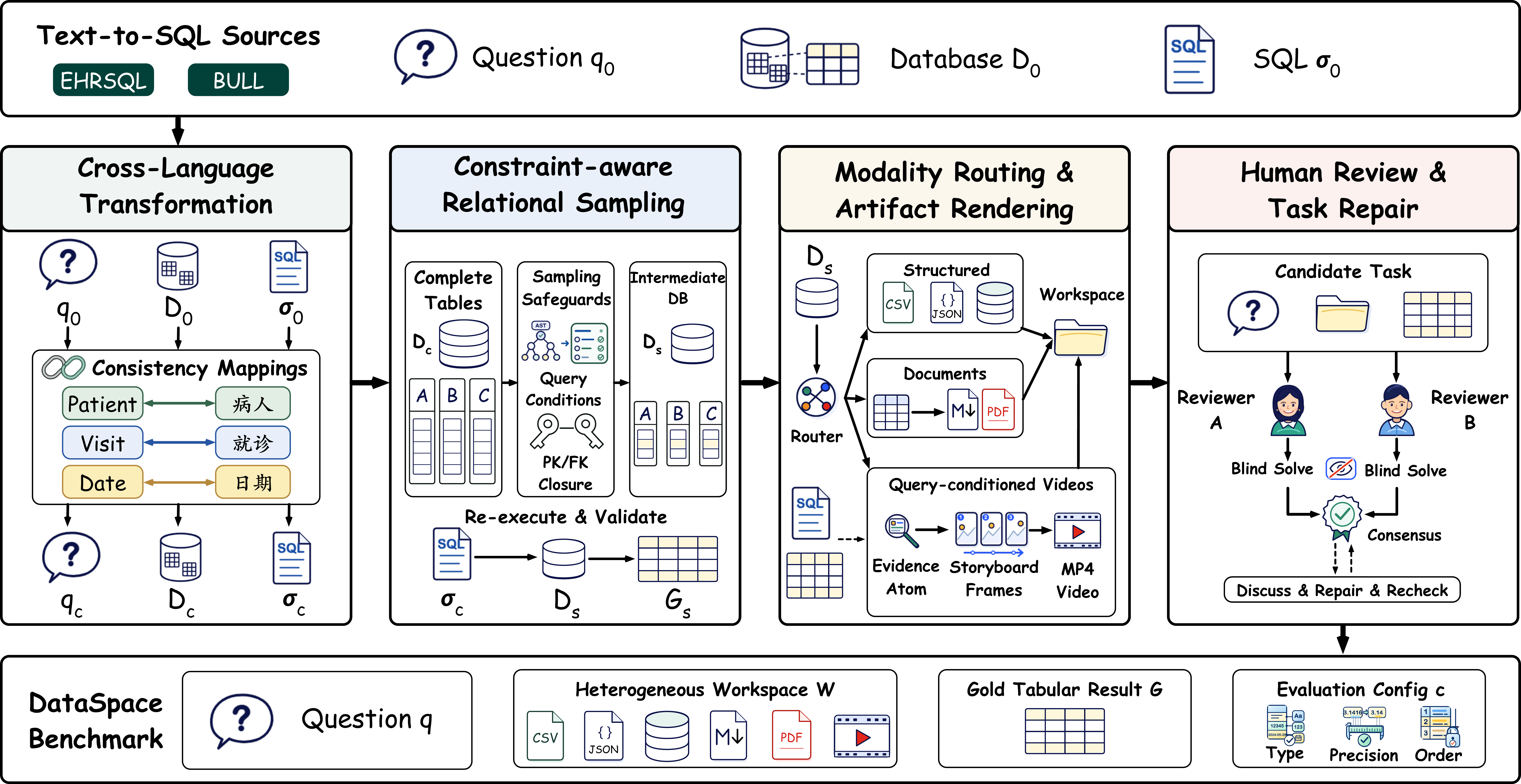}
    \caption{Overview of \dataspacebuilder. Text-to-SQL instances pass through
    Cross-Language Transformation, Constraint-Aware Relational Sampling,
    Modality Routing \& Artifact Rendering, and Human Review \& Task Repair
    before being frozen as heterogeneous benchmark records.}
    \label{fig:construction-overview}
\end{figure*}

In this section, we introduce \dataspacebuilder, a four-stage pipeline that
transforms Text-to-SQL instances into reviewed heterogeneous workspace tasks
(Figure~\ref{fig:construction-overview}). Its stages are Cross-Language
Transformation, Constraint-Aware Relational Sampling, Modality Routing \&
Artifact Rendering, and Human Review \& Task Repair. We then present the
evaluation protocol for finalized tasks. Further implementation details,
construction costs, and running examples appear in
Appendix~\ref{app:additional-details}.

\subsection{Cross-Language Transformation}
\label{sec:cross-language}

\noindent\textbf{Source corpora.}
We use EHRSQL~\cite{ehrsql} and BULL~\cite{finsql} as source corpora.
These English Text-to-SQL benchmarks cover clinical and financial analytics,
respectively. Each selected source instance provides a
natural-language question \(q_0\), a relational database \(\database_0\), and
executable SQL \(\sigma_0\).

\noindent\textbf{Joint transformation.}
Both source corpora are English-only at the question--database level, whereas
our task setting varies language across the complete question--workspace pair.
Translating the question or database in isolation can misalign entity names,
predicate values, and executable SQL. We therefore treat cross-language
transformation (CLT) as a joint migration of the question, database state, and
executable workload. For a source tuple
\((\database_0,q_0,\sigma_0)\), we independently choose the target languages
\(\ell_D\) and \(\ell_q\) for the database and question:
\begin{equation}
    (\database_c,q_c,\sigma_c;M)
    =\mathrm{CLT}_{\ell_D,\ell_q}(\database_0,q_0,\sigma_0),
    \label{eq:cross-language-transform}
\end{equation}
where \(\database_c\), \(q_c\), and \(\sigma_c\) are the transformed database,
question, and SQL, respectively, and \(M\) is a materialized
replacement map.

\noindent\textbf{Consistency-aware rewriting.}
To keep the same field or entity consistent wherever it appears, we link
columns connected by a foreign key, a shared name, or substantial value
overlap. Each linked group is translated jointly, while identifiers, codes,
URLs, dates, and numbers remain unchanged. LLM-generated translations are
stored as table-, column-, and cell-level mappings
\(M=\{M_{\mathrm{tab}},M_{\mathrm{col}},M_{\mathrm{val}}\}\). We use \(M\)
to deterministically rewrite the database and SQL and translate the
question with the same terminology.

\noindent\textbf{Validation.}
We retain a transformed tuple only after structural and execution checks.
The rewritten SQL must reference existing translated entities and execute on
\(\database_c\), and its result must satisfy
\(\operatorname{Exec}(\database_c,\sigma_c)\cong
\tau_M(\operatorname{Exec}(\database_0,\sigma_0))\), where \(\tau_M\) applies
the induced value translation to the source execution result.
Execution equivalence verifies database--SQL consistency but cannot establish
that \(q_c\) preserves the original intent; we therefore use an LLM judge to
verify question--SQL semantic alignment. Instances that fail either check are
repaired and re-executed or rejected before relational data sampling.

\subsection{Constraint-Aware Relational Sampling}
\label{sec:data-sampling}

\noindent\textbf{Task-local data diversification.}
Source Text-to-SQL datasets commonly associate many questions with a small
number of shared database states. Directly reusing these states would produce
workspaces with highly repetitive entities, values, and relational
neighborhoods. We instead construct task-local relational instances by
sampling rows while retaining the complete table inventory and schema. The
sampled scale also makes downstream artifact rendering tractable, particularly
when complete tables are transformed into long Markdown or PDF documents.

\noindent\textbf{Sampling safeguards.}
Naively sampling each table independently can remove a condition value, break
a join path, or disconnect related entities. We therefore construct a
lightweight safeguard set \(\mathcal{C}_s\) that combines primary keys,
foreign keys, and known inter-table relationships with join columns,
predicate and boundary values, and target entities extracted from the SQL
AST.

\noindent\textbf{Relationally consistent sampling.}
Given a seeded sampling policy \(\psi\), we retain rows required by
query safeguards and then propagate their key values across schema
relationships. Remaining rows are selected according to the table-level
sampling budget, including rows from tables not referenced by the source SQL.

\noindent\textbf{Materialization and re-execution.}
All sampled tables are materialized as a task-local intermediate database:
\begin{equation}
    \database_s =
    \operatorname{Sample}(\database_c;\psi,\mathcal{C}_s),
    \qquad
    \goldrel_s =
    \operatorname{Exec}(\database_s,\sigma_c).
    \label{eq:sample-and-execute}
\end{equation}
The query result \(\goldrel_s\) becomes the candidate reference. It is not
required to equal the source result; sampling may change
entities, aggregates, rankings, or result cardinality. A sample is accepted
only if the SQL executes successfully, required relationships remain valid,
and the result has not become unintentionally empty or semantically degenerate.
Otherwise, the database is resampled, repaired, or rejected.

\subsection{Modality Routing \& Artifact Rendering}
\label{sec:artifact-rendering}

\noindent\textbf{Query-independent base routing.}
The intermediate database fixes the relational content of a task; this stage
changes how agents encounter that content. We first apply a seeded rule-based
policy \(\pi_r\), with fixed seed \(z_r\), that assigns each sampled table to
one or more compatible renderers and materializes the base workspace
\(\workspace_{\mathrm{base}}\) from \(\database_s\). The policy considers
schema properties, renderer compatibility, and batch-level modality coverage
using only sampled-table metadata. Its base renderers are CSV, JSON, SQLite,
Markdown, and PDF. Video is introduced
separately as a task-level, query-conditioned augmentation because its
construction may depend on the question, executable SQL, and candidate answer.
Every sampled table receives a base representation, while the number and sizes
of the resulting artifacts follow from the source table inventory, sampled
contents, and applicable renderers.

\noindent\textbf{Structured artifact rendering.}
CSV and record-oriented JSON expose individual sampled tables, whereas a
task-local SQLite artifact can retain several related tables and their schema.
The renderers preserve headers, cell values, nulls, and duplicate rows; the
SQLite renderer additionally preserves declared column types and key
relationships. Each output is parsed back into a canonical relation and
compared with the rows assigned to that renderer.

\noindent\textbf{Fact-grounded data documents.}
For a routed table \(R\), let \(X_R=\database_s[R]\) denote its complete
sampled contents. The LLM-produced plan
\(p_R=\operatorname{PlanDoc}(X_R)\) specifies a document style,
record-identifying columns, semantic attribute groups, and bounded row
batches. Document generation then follows
\begin{equation}
    \begin{aligned}
        \mathcal{B}_R&=\operatorname{BuildBlocks}(X_R;p_R),\\
        d_R&=\operatorname{Assemble}
        (\operatorname{DocGen}(\mathcal{B}_R;p_R)).
    \end{aligned}
    \label{eq:document-generation}
\end{equation}
Each block repeats the identifying columns and supplies an LLM with its exact
field names, types, and values. The generated sections are assembled as
Markdown \(d_R\) and converted to PDF when required.
Figure~\ref{fig:document-running-example} visualizes the alignment from source
cells to generation blocks and document spans. The complete running example
is provided in Appendix~\ref{app:document-running-example}.

\noindent\textbf{Query-conditioned video rendering.}
Following DataMagic~\cite{datamagic}, we render tabular data as data-insight
videos, while conditioning content selection and task integration on the
query.
For selected tasks, the renderer derives typed evidence atoms from the
executable SQL \(\sigma_c\) and sampled tabular result \(\goldrel_s\), then
constructs the video and integrates it into the task:
\begin{equation}
    \begin{aligned}
        \mathcal{E}_v&=\operatorname{Select}
        (\operatorname{AST}(\sigma_c),\goldrel_s),\\
        S_v&=\operatorname{Storyboard}(\mathcal{E}_v),\quad
        v=\operatorname{Render}(S_v,\operatorname{TTS}(S_v)),\\
        (q_r,\workspace)&=\operatorname{Integrate}
        (q_c,\workspace_{\mathrm{base}},v,\mathcal{E}_v).
    \end{aligned}
    \label{eq:video-augmentation}
\end{equation}
The evidence set \(\mathcal{E}_v\) contains filter conditions and result
cells. A filter condition records a column, operator, and value; a result cell
records a row, field, and value. For a multi-step query,
\emph{predicate abstraction} assigns a stable
condition to a video scene and removes it from the explicit question. For a
simple query with a compact result, \emph{answer-evidence rendering}
distributes result cells or an intermediate value across multiple scenes.
We revise the question only when evidence is moved to video.
Figure~\ref{fig:video-running-frames} shows representative frames from both
strategies. Appendix~\ref{app:video-running-example} further traces the
question, evidence atoms, storyboard scenes, rendered video, and resulting
workspace for complete tasks.
For tasks without video augmentation,
\((q_r,\workspace)=(q_c,\workspace_{\mathrm{base}})\).

\subsection{Human Review \& Task Repair}
\label{sec:human-review}

Execution-based checks verify data--SQL consistency, but cannot determine
whether a task is unambiguous or its reference result is correct. We
therefore subject every candidate to blind, independent review by two
reviewers from a panel of 11 domain experts. Each reviewer first solves the
task using only its final question and workspace. After the gold is revealed,
both reviewers verify it and independently author the evaluation configuration
\(c_i\), covering column types, numeric precision, and ordering. Acceptance
requires agreement on the gold and identical canonical configurations. Any
disagreement triggers evidence-based discussion and minimal repair of the
question, workspace, gold, or configuration, followed by independent
rechecking by the same pair. This cycle continues until consensus; unresolved
tasks are removed.
Appendix~\ref{app:human-review-details} gives the full protocol, and
Figure~\ref{fig:review-system-interface} shows the review interface.

\subsection{Evaluation Protocol}
\label{sec:evaluation}

\noindent\textbf{Task-specific semantics.}
After human review, each task freezes a configuration
\(c_i=(o_i,\{\nu_{ij}\}_{j=1}^{d_i})\), where \(d_i\) is the number of
reference columns, \(o_i\) indicates whether row order is required by the
question, and \(\nu_{ij}\) is the canonicalization rule for reference column
\(j\). The rule records a semantic type---text, number, date, datetime, or
Boolean---and, for numeric columns, the required integer, decimal-place, or
significant-digit precision and any percentage convention. It canonicalizes
both reference and predicted values: text is normalized to Unicode NFC and
trimmed; numbers are parsed as decimals under the configured precision and
unit; and dates, datetimes, Booleans, and nulls are converted to canonical
values. Appendix~\ref{app:evaluation-details} gives the complete protocol,
and Figure~\ref{fig:evaluation-config-example} shows a frozen task
configuration.

\noindent\textbf{Header-invariant joint column alignment.}
Prediction headers are not scored, and predicted columns need not follow the
reference order. Instead, the evaluator searches for a one-to-one mapping
from reference columns to predicted columns. Let \(\Pi_{d_i}\) be the set of
permutations of \(d_i\) columns, and let \(\pi(j)\) denote the predicted
column mapped to reference column \(j\). We use
\(\prediction_i[:,\pi]\) to denote the prediction with its columns reordered
by \(\pi\), and \(\operatorname{Canon}_{c_i}(Y)\) to normalize column \(j\)
of \(Y\) with \(\nu_{ij}\). A mapping is invalid if any predicted cell cannot
be interpreted under its target rule.

\noindent\textbf{Answer equivalence.}
Tabular results with different shapes are unequal. Otherwise,
\(\operatorname{Rows}_{o}(Y)\) returns the ordered row sequence when \(o=1\),
and the unordered row multiset when \(o=0\); the latter preserves duplicate
multiplicities. Define the canonical row view as
\(\mathcal{V}_i(Y)=\operatorname{Rows}_{o_i}
(\operatorname{Canon}_{c_i}(Y))\). The equivalence criterion introduced in
Equation~\ref{eq:task-correctness} is then
\begin{equation}
\prediction_i\equiv_{c_i}\goldrel_i
\iff \exists\pi\in\Pi_{d_i}:\quad
\mathcal{V}_i(\prediction_i[:,\pi])=\mathcal{V}_i(\goldrel_i).
\label{eq:relation-equivalence}
\end{equation}
Column alignment is evaluated over the full result, so a valid mapping must
preserve the association among values within every row. Tasks requesting a
ranking or another semantically ordered output are compared as row sequences;
all other tasks are compared as unordered row multisets.

\noindent\textbf{Aggregate metric.}
For \(N\) tasks, the official metric is Task Accuracy,
\begin{equation}
    \mathrm{Acc}
    =\frac{1}{N}\sum_{i=1}^{N}
      \mathbf{1}\!\left[\prediction_i\equiv_{c_i}\goldrel_i\right].
    \label{eq:task-accuracy}
\end{equation}

\section{Benchmark Statistics and Analysis}
\label{sec:benchmark-analysis}

\subsection{Composition and Scale}
\label{sec:benchmark-composition}

\dataspace contains 410 tasks: 363 (88.5\%) originate from BULL and 47
(11.5\%) from EHRSQL. The benchmark covers fund (158 tasks; 38.5\%), stock
(120; 29.3\%), macroeconomic (85; 20.7\%), and healthcare analytics (47;
11.5\%). Of these tasks, 265 (64.6\%) are cross-language and 145 (35.4\%) are
single-language. Their workspaces span 25,384 PDF pages, 55.36 million PDF
and Markdown characters, and 5.49 hours of video. Reference answers contain
126,409 rows; the largest has 12,962 rows, the widest has six columns, and 92
tasks require row order to be preserved. Table~\ref{tab:benchmark-scale}
summarizes the resulting scale.

\begin{table}[t]
    \caption{Workspace scale of \dataspace (410 tasks).}
    \label{tab:benchmark-scale}
    \centering
    \footnotesize
    \setlength{\tabcolsep}{1.5pt}
    \renewcommand{\arraystretch}{0.95}
    \setlength{\aboverulesep}{0.25ex}
    \setlength{\belowrulesep}{0.40ex}

    \begin{tabular}{@{}p{0.19\columnwidth}p{0.32\columnwidth}
                        p{0.21\columnwidth}p{0.23\columnwidth}@{}}
        \toprule
        \textbf{Statistic} & \textbf{Overall} & \textbf{Median}
            & \textbf{P90 / range} \\
        \midrule
        Artifacts & 7,439 files & 20/task & P90: 23; 5--26 \\
        Storage & 15.01 GB & 31.26 MB/task & P90: 74.18 MB \\
        PDF pages & 1,088 PDFs; 25,384 pp. & 22 pages/PDF & P90: 36 \\
        PDF text & 28.48M characters
            & 23,232/PDF & P90: 39,522 \\
        Markdown & 875 files; 26.88M chars
            & 26,190/file & P90: 49,238 \\
        Video & 189 videos; 5.49 h
            & 103.8 s/video & 39.1--158.3 s \\
        \bottomrule
    \end{tabular}
\end{table}

\subsection{Evidence Requirements}
\label{sec:benchmark-evidence}

We annotate one minimal, independently sufficient solution path per task to
distinguish available artifacts from required evidence. We group CSV/JSON,
SQLite, Markdown/PDF, and video as structured-file, database, document, and
video carriers.

\begin{figure}[t]
    \centering
    \begin{subfigure}[t]{\columnwidth}
        \centering
        \includegraphics[width=\linewidth]{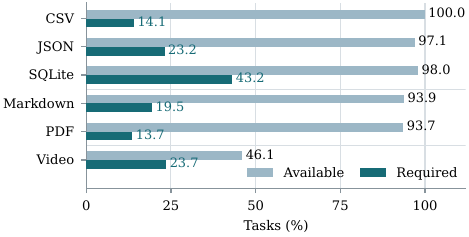}
        \caption{Available and required modalities.}
        \label{fig:evidence-availability}
    \end{subfigure}
    \vspace{-2pt}
    \begin{subfigure}[t]{\columnwidth}
        \centering
        \includegraphics[width=\linewidth]{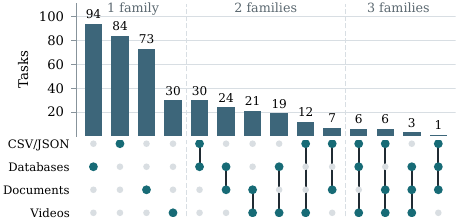}
        \caption{Required carrier-family intersections.}
        \label{fig:evidence-carriers}
    \end{subfigure}
    \caption{Workspace availability and annotated solution evidence.
    Available denotes presence in the workspace, whereas required denotes use
    by the verified solution path.}
    \label{fig:evidence-requirements}
\end{figure}

\noindent\textbf{Available versus required evidence.}
Figure~\ref{fig:evidence-availability} shows that CSV occurs in every
workspace, while JSON, SQLite, Markdown, and PDF each occur in more than
93\% of tasks, yet the verified paths use CSV in only 58 tasks and SQLite in
177. Long documents provide required evidence in 135 tasks; among 189
video-enabled workspaces, 97 require video.

\noindent\textbf{Cross-artifact composition.}
The verified paths use one artifact modality for 276 tasks (67.3\%), two for
115 (28.0\%), and three for 19 (4.6\%); thus, 134 tasks (32.7\%) combine
multiple modalities. At the carrier-family level in
Figure~\ref{fig:evidence-carriers}, 129 tasks (31.5\%) span multiple carrier
families, and 202 (49.3\%) require a document or video.

\subsection{Analytical Operations}
\label{sec:benchmark-operations}

\begin{figure}[t]
    \centering
    \includegraphics[width=\columnwidth]{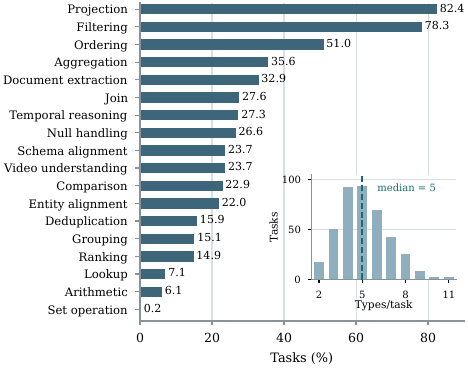}
    \caption{Analytical-operation coverage and compositional breadth. Bars
    show task-level prevalence; the inset shows distinct operation types per
    task.}
    \label{fig:operation-coverage}
\end{figure}

\noindent\textbf{Operation coverage.}
As shown in Figure~\ref{fig:operation-coverage}, projection and filtering are
required by 338 (82.4\%) and 321 tasks (78.3\%), respectively. The benchmark
also exercises ordering in 209 tasks (51.0\%), aggregation in 146 (35.6\%),
and joins in 113 (27.6\%). Cross-artifact grounding includes document and
video understanding in 135 and 97 tasks, and schema and entity alignment in
97 and 90.

\noindent\textbf{Compositional breadth.}
A task combines between two and eleven nontrivial operation types, with a
median of five and a 90th percentile of seven; 248 tasks (60.5\%) combine at
least five types.

\section{Experiments}
\label{sec:experiments}

\subsection{Experimental Setup}
\label{sec:experimental-setup}

We evaluate all 410 tasks under two complementary controls. For the
\emph{backbone comparison}, we implement \dataspaceagent, a lightweight agent
that follows the ReAct~\cite{react} paradigm and exposes a minimal set of
task-agnostic tools. We fix this agent and vary only the multimodal backbone,
testing six models released between April and July 2026:
Grok~4.5~\cite{grok45}, GPT-5.6~Sol~\cite{gpt56sol},
Kimi~K3~\cite{kimik3}, MiMo-V2.5~\cite{mimov25}, Claude
Sonnet~5~\cite{claudesonnet5}, and MiniMax~M3~\cite{minimaxm3}. In the
\emph{harness comparison}, we fix MiMo-V2.5 and compare \dataspaceagent,
Smolagents~\cite{smolagents}, Codex~\cite{codexcli}, Claude
Code~\cite{claudecode}, and Grok Build~\cite{grokbuild} while retaining each
harness's native planning, tool-use, and context management. Each
backbone--harness pair instantiates a data agent. We omit specialized systems
that cannot be evaluated faithfully under this protocol, such as
DeepAnalyze~\cite{deepanalyze}, whose native interface does not cover video
inputs or our tabular-output contract, and AgenticData~\cite{agenticdata},
whose original implementation is unavailable.
Model and harness configurations, together with the broader compatibility
assessment, are provided in Appendix~\ref{app:experimental-details}.

\noindent\textbf{Execution protocol.}
Each run receives the task question and its complete workspace. In the
backbone comparison, every task is limited to 60 model turns, 50 tool actions,
and 1,800 seconds, with 4 CPUs and 16\,GiB of memory. The harness comparison
uses the same 1,800-second deadline but does not impose a shared action limit
that would override a harness's native control loop.

\noindent\textbf{Inference and scoring.}
All backbone calls use the provider-default reasoning configuration through
Vercel AI Gateway~\cite{vercelaigateway}, with a maximum of 32,768 output
tokens per call. The official evaluator computes Task Accuracy using the
protocol in Section~\ref{sec:evaluation}. Missing
predictions, invalid outputs, runtime failures, and exhausted budgets count as
incorrect. We additionally record token usage, API cost, tool actions, and
wall-clock latency as efficiency diagnostics.

\subsection{Overall Effectiveness}
\label{sec:overall-effectiveness}

Table~\ref{tab:overall-effectiveness} reports Task Accuracy for both controlled
comparisons.

\begin{table}[t]
\caption{Overall effectiveness in controlled comparisons. Correct is out of
410; snapshot denotes backbone release month or harness version. Best in each
block is bold.}
\label{tab:overall-effectiveness}
\centering
\small
\setlength{\tabcolsep}{2.5pt}
\renewcommand{\arraystretch}{0.96}
\begin{tabular*}{\columnwidth}{@{\extracolsep{\fill}}lrrr@{}}
\toprule
\textbf{Method} & \textbf{Snapshot} & \textbf{Correct} &
\textbf{Acc. (\%)} \\
\midrule
\multicolumn{4}{@{}l}{\textit{Backbones (\dataspaceagent fixed)}} \\
Grok 4.5~\cite{grok45} &
    2026-07 & \textbf{272} & \textbf{66.34} \\
GPT-5.6 Sol~\cite{gpt56sol} &
    2026-07 & 265 & 64.63 \\
Kimi K3~\cite{kimik3} &
    2026-07 & 219 & 53.41 \\
MiMo-V2.5~\cite{mimov25} &
    2026-04 & 161 & 39.27 \\
Claude Sonnet 5~\cite{claudesonnet5} &
    2026-06 & 135 & 32.93 \\
MiniMax M3~\cite{minimaxm3} &
    2026-06 & 117 & 28.54 \\
\midrule
\multicolumn{4}{@{}l}{\textit{Harnesses
    (MiMo-V2.5~\cite{mimov25} fixed)}} \\
Grok Build~\cite{grokbuild} &
    v0.2.106 & \textbf{190} & \textbf{46.34} \\
Claude Code~\cite{claudecode} &
    v2.1.217 & 183 & 44.63 \\
\dataspaceagent \textit{(ours)} &
    \nomark & 161 & 39.27 \\
Codex~\cite{codexcli} &
    v0.145.0 & 143 & 34.88 \\
Smolagents~\cite{smolagents} &
    v1.26.0 & 127 & 30.98 \\
\bottomrule
\end{tabular*}
\end{table}

\noindent\textbf{Backbone comparison.}
With \dataspaceagent fixed, Grok~4.5 achieves the highest observed accuracy of
66.34\%, followed by GPT-5.6~Sol at 64.63\%; their totals differ by only seven
correct tasks. Kimi~K3 reaches 53.41\%, while the remaining three
backbones remain below 40\%. The 37.80-point range between the strongest and
weakest backbone shows substantial separation, yet even the best result
solves only about two thirds of the benchmark. Across the six models, 56 tasks
are solved by all models, whereas 76 are solved by none; their oracle union
solves 334 tasks (81.46\%). Thus, the benchmark contains both a shared hard
core and model-specific successes that are hidden by aggregate ranking alone.

\noindent\textbf{Harness comparison.}
Fixing MiMo-V2.5, Grok Build obtains 46.34\% and Claude Code 44.63\%, compared
with 39.27\% for \dataspaceagent, 34.88\% for Codex, and 30.98\% for
Smolagents. The resulting 15.36-point spread demonstrates that the harness
substantially affects end-to-end task completion even when the backbone is
held constant.

\findingbox{1}{\dataspace remains unsaturated: the best controlled backbone
reaches 66.34\% Task Accuracy, while 76 tasks are missed by all six backbones.
Harness design also substantially affects agent performance: with MiMo-V2.5
fixed, accuracy ranges from 30.98\% to 46.34\%.}

\subsection{Efficiency and Trade-offs}
\label{sec:efficiency}

\begin{figure}[t]
    \centering
    \includegraphics[width=\columnwidth]{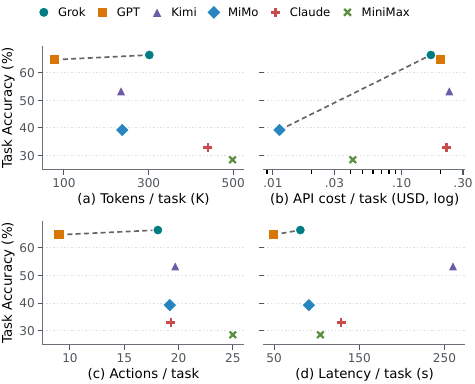}
    \caption{Backbone accuracy--efficiency trade-offs with \dataspaceagent
    fixed. Dashed lines connect Pareto-efficient points (higher accuracy,
    lower resource use).}
    \label{fig:backbone-efficiency}
\end{figure}

We fix \dataspaceagent to compare backbone efficiency under a common agent
design. Figure~\ref{fig:backbone-efficiency} relates Task Accuracy to mean
token usage, API cost, tool actions, and wall-clock latency. GPT-5.6 Sol
reaches 64.63\% accuracy, only 1.71 points below Grok 4.5, while using 74.2\%
fewer tokens, 50.3\% fewer actions, and 39.2\% less wall-clock time per task.
These two models form the token-, action-, and latency-based Pareto frontiers:
GPT provides the most compact near-top solution, while Grok trades additional
resources for the highest accuracy. The cost frontier differs: MiMo-V2.5 costs
only \$0.011 per task at 39.27\% accuracy, whereas Grok reaches 66.34\% at
\$0.169 and GPT costs \$0.200 per task. Complete statistics, including medians
and tail latency, appear in Appendix~\ref{app:backbone-efficiency}.

\findingbox{2}{Near-top accuracy need not require long trajectories:
GPT-5.6 Sol trails Grok 4.5 by 1.71 points while using 74.2\% fewer tokens,
50.3\% fewer actions, and 39.2\% less latency. Monetary efficiency follows a
different frontier, with MiMo-V2.5 providing the lowest-cost operating point
at \$0.011 per task.}

\subsection{Performance across Task Characteristics}
\label{sec:performance-variation}

\begin{figure}[t]
    \centering
    \includegraphics[width=\columnwidth]{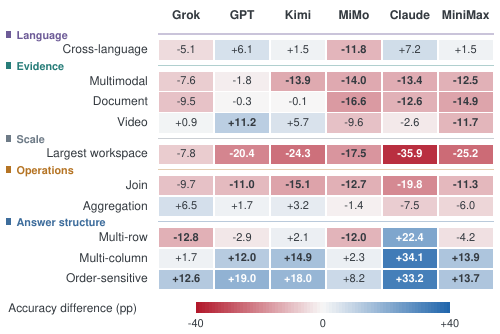}
    \caption{Task Accuracy differences with \dataspaceagent fixed. Cells
    report percentage-point changes from each row's reference group; positive
    values favor the named row.}
    \label{fig:task-characteristic-sensitivity}
\end{figure}

We stratify the six \dataspaceagent runs by the annotations in
Section~\ref{sec:benchmark-analysis}. Figure~\ref{fig:task-characteristic-sensitivity}
reports Task Accuracy gaps across overlapping task characteristics; these are
descriptive rather than causal. Sample sizes and raw accuracies appear in
Appendix~\ref{app:performance-subgroups}.

\noindent\textbf{Language and workspace scale.}
Cross-language performance varies by backbone: relative to single-language
tasks, MiMo-V2.5 declines by 11.8 points and Grok 4.5 by 5.1, whereas GPT-5.6
Sol and Claude Sonnet 5 improve by 6.1 and 7.2 points. The largest workspace
quartile underperforms the smallest for all six models, but accuracy across
the four quartiles is not monotonic. At the task level, workspace bytes have
only a weak negative Spearman correlation (\(-0.186\)) with the number of
models that solve a task.

\noindent\textbf{Evidence composition.}
Multimodal tasks underperform single-modal tasks for every backbone by
1.8--14.0 points. Modality presence alone has a less uniform effect: required
document evidence is nearly neutral for GPT and Kimi but substantially
reduces accuracy for Grok, MiMo, Claude, and MiniMax. Required video evidence
helps GPT and Kimi but hurts MiMo, Claude, and MiniMax. These contrasts locate
the consistent challenge in cross-modality integration rather than in any
single modality.

\noindent\textbf{Relational and answer requirements.}
Join requirements reduce accuracy for all backbones by 9.7--19.8 points,
whereas aggregation has mixed effects. Answer shape is similarly
non-monotonic: multi-row answers hurt Grok and MiMo but improve Claude, while
multi-column and order-sensitive groups are not less accurate overall. Answer
size and order therefore show no uniform negative association with accuracy;
the analytical path and task composition remain more informative.

\findingbox{3}{Cross-modality integration and joins are the most consistent
sources of degradation: multimodal tasks underperform single-modal tasks for
every backbone by 1.8--14.0 points, and joins reduce accuracy by 9.7--19.8 points.
Language configuration and answer shape instead exhibit backbone-dependent
effects.}

\subsection{Failure Analysis}
\label{sec:failure-analysis}

We conduct a trace-level error analysis of 136 failures from Grok 4.5, the
strongest controlled backbone.

\begin{figure}[!t]
    \centering
    \includegraphics[width=\columnwidth]{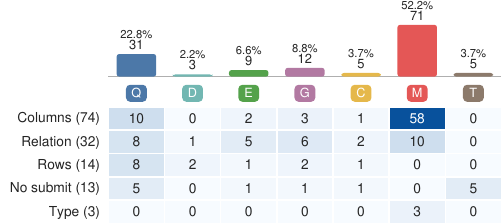}
    \caption{Error analysis of 136 Grok 4.5 failures.
    Bars show primary causes; the matrix decomposes evaluator symptoms.
    Q/D/E/G/C/M/T denote intent, discovery, extraction, grounding,
    computation, materialization, and termination.}
    \label{fig:failure-analysis}
\end{figure}

\noindent\textbf{Where failures originate.}
Answer materialization is the largest category, accounting for 71/136
(52.2\%) failures, followed by task specification and intent with 31/136
(22.8\%). At the subtype level, 60 materialization failures add or omit
columns after the needed internal result is available, while 17 intent
failures misformulate the requested output or row grain. These two
distinct routes to an incorrect answer schema comprise 77/136 (56.6\%) of
audited failures. By comparison, only three failures arise from selecting the
wrong evidence source; extraction and semantic grounding together account for
21, showing that locating an artifact does not ensure that its values are
recovered and aligned correctly.

\noindent\textbf{Symptoms are not diagnoses.}
Figure~\ref{fig:failure-analysis} shows that the same evaluator outcome can
arise at different stages. Of 74 audited column-count mismatches,
58 originate in materialization, while the remainder begin in task intent,
extraction, grounding, or computation. Conversely, only 5 of the 13
no-submission outcomes are pure execution-control failures; the other eight
follow an earlier persistent interpretation, extraction, grounding, or
computation error. Aggregate output symptoms therefore obscure the upstream
intervention required to correct a trajectory. The complete subtype taxonomy
and counts appear in Appendix~\ref{app:failure-analysis}.

\findingbox{4}{Harnesses must faithfully materialize the exact requested output,
rather than merely find and correctly compute the relevant values:
target-result misunderstanding and faulty
column projection account for 56.6\% of Grok 4.5's audited
failures, while only 5 of 13 no-submission outcomes are pure termination
failures.}

\section{Conclusion}
\label{sec:conclusion}

We introduced \dataspace, a 410-task benchmark for verifiable analytics with
complete tabular outputs over cross-language, heterogeneous workspaces.
\dataspacebuilder transforms
executable Text-to-SQL instances into multimodal tasks with expert review,
paired with semantics-aware tabular-result evaluation. Across six backbones and five
harnesses, the best accuracy reaches 66.34\%, while multimodal evidence
integration and joins remain key bottlenecks. \dataspace establishes a
rigorous test bed for advancing reliable data agents over heterogeneous
workspaces.

\clearpage
\bibliographystyle{ACM-Reference-Format}
\bibliography{references}

@inproceedings{nvbench2,
  author       = {Tianqi Luo and
                  Chuhan Huang and
                  Leixian Shen and
                  Boyan Li and
                  Shuyu Shen and
                  Wei Zeng and
                  Nan Tang and
                  Yuyu Luo},
  title        = {nvBench 2.0: Resolving Ambiguity in Text-to-Visualization through
                  Stepwise Reasoning},
  booktitle    = {NeurIPS},
  year         = {2025}
}

@article{elliesql,
  author       = {Yizhang Zhu and
                  Runzhi Jiang and
                  Boyan Li and
                  Nan Tang and
                  Yuyu Luo},
  title        = {EllieSQL: Cost-Efficient Text-to-SQL with Complexity-Aware Routing},
  journal      = {CoRR},
  volume       = {abs/2503.22402},
  year         = {2025}
}

@article{texttosqllmsurvey,
  author       = {Xinyu Liu and
                  Shuyu Shen and
                  Boyan Li and
                  Peixian Ma and
                  Runzhi Jiang and
                  Yuxin Zhang and
                  Ju Fan and
                  Guoliang Li and
                  Nan Tang and
                  Yuyu Luo},
  title        = {A Survey of Text-to-SQL in the Era of LLMs: Where Are We, and Where
                  Are We Going?},
  journal      = {{IEEE} Trans. Knowl. Data Eng.},
  volume       = {37},
  number       = {10},
  pages        = {5735--5754},
  year         = {2025}
}

@inproceedings{nl2sqlbugs,
  author       = {Xinyu Liu and
                  Shuyu Shen and
                  Boyan Li and
                  Nan Tang and
                  Yuyu Luo},
  title        = {NL2SQL-BUGs: {A} Benchmark for Detecting Semantic Errors in {NL2SQL}
                  Translation},
  booktitle    = {{KDD} {(2)}},
  pages        = {5662--5673},
  publisher    = {{ACM}},
  year         = {2025}
}

@article{dataagentsurvey,
  author       = {Yizhang Zhu and
                  Liangwei Wang and
                  Chenyu Yang and
                  Xiaotian Lin and
                  Boyan Li and
                  Wei Zhou and
                  Xinyu Liu and
                  Zhangyang Peng and
                  Tianqi Luo and
                  Yu Li and
                  Chengliang Chai and
                  Chong Chen and
                  Shimin Di and
                  Ju Fan and
                  Ji Sun and
                  Nan Tang and
                  Fugee Tsung and
                  Jiannan Wang and
                  Chenglin Wu and
                  Yanwei Xu and
                  Shaolei Zhang and
                  Yong Zhang and
                  Xuanhe Zhou and
                  Guoliang Li and
                  Yuyu Luo},
  title        = {A Survey of Data Agents: Emerging Paradigm or Overstated Hype?},
  journal      = {CoRR},
  volume       = {abs/2510.23587},
  year         = {2025}
}

@article{deepvis,
  author       = {Zhihao Shuai and
                  Boyan Li and
                  Siyu Yan and
                  Yuyu Luo and
                  Weikai Yang},
  title        = {DeepVIS: Bridging Natural Language and Data Visualization Through
                  Step-Wise Reasoning},
  journal      = {{IEEE} Trans. Vis. Comput. Graph.},
  volume       = {32},
  number       = {1},
  pages        = {868--878},
  year         = {2026}
}

@article{liang2026long,
  author       = {Zhuowen Liang and
                  Xiaotian Lin and
                  Zhengxuan Zhang and
                  Yuyu Luo and
                  Haixun Wang and
                  Nan Tang},
  title        = {Long-Document {QA} with Chain-of-Structured-Thought and Fine-Tuned
                  SLMs},
  journal      = {CoRR},
  volume       = {abs/2603.29232},
  year         = {2026}
}

@article{zhang2025datamosaic,
  author       = {Zhengxuan Zhang and
                  Zhuowen Liang and
                  Yin Wu and
                  Teng Lin and
                  Yuyu Luo and
                  Nan Tang},
  title        = {DataMosaic: Explainable and Verifiable Multi-Modal Data Analytics
                  through Extract-Reason-Verify},
  journal      = {CoRR},
  volume       = {abs/2504.10036},
  year         = {2025}
}

@article{dawnnl2sql,
  author       = {Boyan Li and
                  Yuyu Luo and
                  Chengliang Chai and
                  Guoliang Li and
                  Nan Tang},
  title        = {The Dawn of Natural Language to {SQL:} Are We Fully Ready? [Experiment,
                  Analysis {\&} Benchmark {]}},
  journal      = {Proc. {VLDB} Endow.},
  volume       = {17},
  number       = {11},
  pages        = {3318--3331},
  year         = {2024}
}

@inproceedings{alphasql,
  author       = {Boyan Li and
                  Jiayi Zhang and
                  Ju Fan and
                  Yanwei Xu and
                  Chong Chen and
                  Nan Tang and
                  Yuyu Luo},
  title        = {Alpha-SQL: Zero-Shot Text-to-SQL using Monte Carlo Tree Search},
  booktitle    = {{ICML}},
  series       = {Proceedings of Machine Learning Research},
  volume       = {267},
  publisher    = {{PMLR} / OpenReview.net},
  year         = {2025}
}

@article{deepeyesql,
  author       = {Boyan Li and
                  Chong Chen and
                  Zhujun Xue and
                  Yinan Mei and
                  Yuyu Luo},
  title        = {DeepEye-SQL: {A} Software-Engineering-Inspired Text-to-SQL Framework},
  journal      = {CoRR},
  volume       = {abs/2510.17586},
  year         = {2025}
}

@inproceedings{dpc,
  author       = {Boyan Li and
                  Ou Ocean Kun Hei and
                  Yue Yu and
                  Yuyu Luo},
  title        = {{DPC:} Training-Free Text-to-SQL Candidate Selection via Dual-Paradigm
                  Consistency},
  booktitle    = {{ACL} {(1)}},
  pages        = {6897--6913},
  publisher    = {Association for Computational Linguistics},
  year         = {2026}
}

@inproceedings{rose,
  author       = {Wenqi Pei and
                  Shizheng Hou and
                  Boyan Li and
                  Chen Han and
                  Zhichao Shi and
                  Yuyu Luo},
  title        = {{ROSE:} An Intent-Centered Evaluation Metric for {NL2SQL}},
  booktitle    = {{ACL} {(1)}},
  pages        = {5682--5709},
  publisher    = {Association for Computational Linguistics},
  year         = {2026}
}

@article{sqlconductor,
  author       = {Yizhang Zhu and
                  Zhangyang Peng and
                  Boyan Li and
                  Yuyu Luo},
  title        = {SQLConductor: Search-to-Policy Learning for Step-wise Text-to-SQL
                  Orchestration},
  journal      = {CoRR},
  volume       = {abs/2606.23537},
  year         = {2026}
}

@article{nl2sqlrewriter,
  author       = {Peixian Ma and
                  Boyan Li and
                  Runzhi Jiang and
                  Ju Fan and
                  Nan Tang and
                  Yuyu Luo},
  title        = {A Plug-and-Play Natural Language Rewriter for Natural Language to
                  {SQL}},
  journal      = {CoRR},
  volume       = {abs/2412.17068},
  year         = {2024}
}

@inproceedings{spider,
  author       = {Tao Yu and
                  Rui Zhang and
                  Kai Yang and
                  Michihiro Yasunaga and
                  Dongxu Wang and
                  Zifan Li and
                  James Ma and
                  Irene Li and
                  Qingning Yao and
                  Shanelle Roman and
                  Zilin Zhang and
                  Dragomir R. Radev},
  title        = {Spider: {A} Large-Scale Human-Labeled Dataset for Complex and Cross-Domain
                  Semantic Parsing and Text-to-SQL Task},
  booktitle    = {{EMNLP}},
  pages        = {3911--3921},
  publisher    = {Association for Computational Linguistics},
  year         = {2018}
}

@inproceedings{wikitablequestions,
  author       = {Panupong Pasupat and
                  Percy Liang},
  title        = {Compositional Semantic Parsing on Semi-Structured Tables},
  booktitle    = {{ACL} {(1)}},
  pages        = {1470--1480},
  publisher    = {The Association for Computer Linguistics},
  year         = {2015}
}

@inproceedings{bird,
  author       = {Jinyang Li and
                  Binyuan Hui and
                  Ge Qu and
                  Jiaxi Yang and
                  Binhua Li and
                  Bowen Li and
                  Bailin Wang and
                  Bowen Qin and
                  Ruiying Geng and
                  Nan Huo and
                  Xuanhe Zhou and
                  Chenhao Ma and
                  Guoliang Li and
                  Kevin Chen{-}Chuan Chang and
                  Fei Huang and
                  Reynold Cheng and
                  Yongbin Li},
  title        = {Can {LLM} Already Serve as {A} Database Interface? {A} BIg Bench for
                  Large-Scale Database Grounded Text-to-SQLs},
  booktitle    = {NeurIPS},
  year         = {2023}
}

@article{ehrsql,
  author       = {Gyubok Lee and
                  Hyeonji Hwang and
                  Seongsu Bae and
                  Yeonsu Kwon and
                  Woncheol Shin and
                  Seongjun Yang and
                  Minjoon Seo and
                  Jong{-}Yeup Kim and
                  Edward Choi},
  title        = {{EHRSQL:} {A} Practical Text-to-SQL Benchmark for Electronic Health
                  Records},
  journal      = {CoRR},
  volume       = {abs/2301.07695},
  year         = {2023}
}

@inproceedings{finsql,
  author       = {Chao Zhang and
                  Yuren Mao and
                  Yijiang Fan and
                  Yu Mi and
                  Yunjun Gao and
                  Lu Chen and
                  Dongfang Lou and
                  Jinshu Lin},
  title        = {FinSQL: Model-Agnostic LLMs-based Text-to-SQL Framework for Financial
                  Analysis},
  booktitle    = {{SIGMOD} Conference Companion},
  pages        = {93--105},
  publisher    = {{ACM}},
  year         = {2024}
}

@misc{kddcup2026dataagents,
  author       = {{KDD Cup 2026}},
  title        = {{KDD Cup 2026}: Data Agents for Complex Data Analysis},
  year         = {2026},
  url          = {https://dataagent.top/},
  note         = {Accessed July 27, 2026}
}

@inproceedings{spider2,
  author       = {Fangyu Lei and
                  Jixuan Chen and
                  Yuxiao Ye and
                  Ruisheng Cao and
                  Dongchan Shin and
                  Hongjin Su and
                  Zhaoqing Suo and
                  Hongcheng Gao and
                  Wenjing Hu and
                  Pengcheng Yin and
                  Victor Zhong and
                  Caiming Xiong and
                  Ruoxi Sun and
                  Qian Liu and
                  Sida Wang and
                  Tao Yu},
  title        = {Spider 2.0: Evaluating Language Models on Real-World Enterprise Text-to-SQL
                  Workflows},
  booktitle    = {{ICLR}},
  publisher    = {OpenReview.net},
  year         = {2025}
}

@inproceedings{hotpotqa,
  author       = {Zhilin Yang and
                  Peng Qi and
                  Saizheng Zhang and
                  Yoshua Bengio and
                  William W. Cohen and
                  Ruslan Salakhutdinov and
                  Christopher D. Manning},
  title        = {HotpotQA: {A} Dataset for Diverse, Explainable Multi-hop Question
                  Answering},
  booktitle    = {{EMNLP}},
  pages        = {2369--2380},
  publisher    = {Association for Computational Linguistics},
  year         = {2018}
}

@article{crag,
  author       = {Xiao Yang and
                  Kai Sun and
                  Hao Xin and
                  Yushi Sun and
                  Nikita Bhalla and
                  Xiangsen Chen and
                  Sajal Choudhary and
                  Rongze Daniel Gui and
                  Ziran Will Jiang and
                  Ziyu Jiang and
                  Lingkun Kong and
                  Brian Moran and
                  Jiaqi Wang and
                  Yifan Ethan Xu and
                  An Yan and
                  Chenyu Yang and
                  Eting Yuan and
                  Hanwen Zha and
                  Nan Tang and
                  Lei Chen and
                  Nicolas Scheffer and
                  Yue Liu and
                  Nirav Shah and
                  Rakesh Wanga and
                  Anuj Kumar and
                  Wen{-}tau Yih and
                  Xin Luna Dong},
  title        = {{CRAG} - Comprehensive {RAG} Benchmark},
  journal      = {CoRR},
  volume       = {abs/2406.04744},
  year         = {2024}
}

@article{financebench,
  author       = {Pranab Islam and
                  Anand Kannappan and
                  Douwe Kiela and
                  Rebecca Qian and
                  Nino Scherrer and
                  Bertie Vidgen},
  title        = {FinanceBench: {A} New Benchmark for Financial Question Answering},
  journal      = {CoRR},
  volume       = {abs/2311.11944},
  year         = {2023}
}

@inproceedings{hybridqa,
  author       = {Wenhu Chen and
                  Hanwen Zha and
                  Zhiyu Chen and
                  Wenhan Xiong and
                  Hong Wang and
                  William Yang Wang},
  title        = {HybridQA: {A} Dataset of Multi-Hop Question Answering over Tabular
                  and Textual Data},
  booktitle    = {{EMNLP} (Findings)},
  series       = {Findings of {ACL}},
  volume       = {{EMNLP} 2020},
  pages        = {1026--1036},
  publisher    = {Association for Computational Linguistics},
  year         = {2020}
}

@inproceedings{multimodalqa,
  author       = {Alon Talmor and
                  Ori Yoran and
                  Amnon Catav and
                  Dan Lahav and
                  Yizhong Wang and
                  Akari Asai and
                  Gabriel Ilharco and
                  Hannaneh Hajishirzi and
                  Jonathan Berant},
  title        = {MultiModalQA: complex question answering over text, tables and images},
  booktitle    = {{ICLR}},
  publisher    = {OpenReview.net},
  year         = {2021}
}

@inproceedings{mmlongbenchdoc,
  author       = {Yubo Ma and
                  Yuhang Zang and
                  Liangyu Chen and
                  Meiqi Chen and
                  Yizhu Jiao and
                  Xinze Li and
                  Xinyuan Lu and
                  Ziyu Liu and
                  Yan Ma and
                  Xiaoyi Dong and
                  Pan Zhang and
                  Liangming Pan and
                  Yu{-}Gang Jiang and
                  Jiaqi Wang and
                  Yixin Cao and
                  Aixin Sun},
  title        = {{MMLONGBENCH-DOC:} Benchmarking Long-context Document Understanding
                  with Visualizations},
  booktitle    = {NeurIPS},
  year         = {2024}
}

@inproceedings{videomme,
  author       = {Chaoyou Fu and
                  Yuhan Dai and
                  Yongdong Luo and
                  Lei Li and
                  Shuhuai Ren and
                  Renrui Zhang and
                  Zihan Wang and
                  Chenyu Zhou and
                  Yunhang Shen and
                  Mengdan Zhang and
                  Peixian Chen and
                  Yanwei Li and
                  Shaohui Lin and
                  Sirui Zhao and
                  Ke Li and
                  Tong Xu and
                  Xiawu Zheng and
                  Enhong Chen and
                  Caifeng Shan and
                  Ran He and
                  Xing Sun},
  title        = {Video-MME: The First-Ever Comprehensive Evaluation Benchmark of Multi-modal
                  LLMs in Video Analysis},
  booktitle    = {{CVPR}},
  pages        = {24108--24118},
  publisher    = {Computer Vision Foundation / {IEEE}},
  year         = {2025}
}

@article{dabstep,
  title={Dabstep: Data agent benchmark for multi-step reasoning},
  author={Egg, Alex and Goyanes, Martin Iglesias and Kingma, Friso and Mora, Andreu and von Werra, Leandro and Wolf, Thomas},
  journal={arXiv preprint arXiv:2506.23719},
  year={2025}
}

@article{kramabench,
  author       = {Eugenie Lai and
                  Gerardo Vitagliano and
                  Ziyu Zhang and
                  Sivaprasad Sudhir and
                  Om Chabra and
                  Anna Zeng and
                  Anton A. Zabreyko and
                  Chenning Li and
                  Ferdi Kossmann and
                  Jialin Ding and
                  Jun Chen and
                  Markos Markakis and
                  Matthew Russo and
                  Weiyang Wang and
                  Ziniu Wu and
                  Michael J. Cafarella and
                  Lei Cao and
                  Samuel Madden and
                  Tim Kraska},
  title        = {KramaBench: {A} Benchmark for {AI} Systems on Data-to-Insight Pipelines
                  over Data Lakes},
  journal      = {CoRR},
  volume       = {abs/2506.06541},
  year         = {2025}
}

@article{longda,
  title={LongDA: Benchmarking LLM Agents for Long-Document Data Analysis},
  author={Li, Yiyang and Zhang, Zheyuan and Ma, Tianyi and Wang, Zehong and Murugesan, Keerthiram and Zhang, Chuxu and Ye, Yanfang},
  journal={arXiv preprint arXiv:2601.02598},
  year={2026}
}

@article{datacross,
  author       = {Ruyi Qi and
                  Zhou Liu and
                  Wentao Zhang},
  title        = {DataCross: {A} Unified Benchmark and Agent Framework for Cross-Modal
                  Heterogeneous Data Analysis},
  journal      = {CoRR},
  volume       = {abs/2601.21403},
  year         = {2026}
}

@article{fdabench,
  author       = {Ziting Wang and
                  Shize Zhang and
                  Haitao Yuan and
                  Jinwei Zhu and
                  Shifu Li and
                  Wei Dong and
                  Gao Cong},
  title        = {FDABench: {A} Benchmark for Data Agents on Analytical Queries over
                  Heterogeneous Data},
  journal      = {CoRR},
  volume       = {abs/2509.02473},
  year         = {2025}
}

@article{dataagentbenchmark,
  title={Can ai agents answer your data questions? a benchmark for data agents},
  author={Ma, Ruiying and Shankar, Shreya and Chen, Ruiqi and Lin, Yiming and Zeighami, Sepanta and Ghosh, Rajoshi and Gupta, Abhinav and Gupta, Anushrut and Gopal, Tanmai and Parameswaran, Aditya G},
  journal={arXiv preprint arXiv:2603.20576},
  year={2026}
}

@article{agenticdatabench,
  title={AgenticDataBench: A Comprehensive Benchmark for Data Agents},
  author={Sun, Zhaoyan and Zhong, Shan and Wen, Daizhou and Han, Jiaxing and Li, Guoliang and Yan, Ying and Zhang, Peng and Su, Yu and Qi, Xiang and Sun, Baolin and others},
  journal={arXiv preprint arXiv:2607.01647},
  year={2026}
}

@inproceedings{react,
  author       = {Shunyu Yao and
                  Jeffrey Zhao and
                  Dian Yu and
                  Nan Du and
                  Izhak Shafran and
                  Karthik R. Narasimhan and
                  Yuan Cao},
  title        = {ReAct: Synergizing Reasoning and Acting in Language Models},
  booktitle    = {{ICLR}},
  publisher    = {OpenReview.net},
  year         = {2023}
}

@misc{vercelaigateway,
  author       = {{Vercel}},
  title        = {{AI Gateway}},
  year         = {2026},
  url          = {https://vercel.com/docs/ai-gateway},
  note         = {Accessed July 27, 2026}
}

@misc{grok45,
  author       = {{SpaceXAI}},
  title        = {Introducing {Grok 4.5}},
  year         = {2026},
  month        = jul,
  url          = {https://x.ai/news/grok-4-5}
}

@misc{gpt56sol,
  author       = {{OpenAI}},
  title        = {{GPT-5.6 Sol}},
  year         = {2026},
  url          = {https://developers.openai.com/api/docs/models/gpt-5.6-sol},
  note         = {Accessed July 27, 2026}
}

@misc{kimik3,
  author       = {{Moonshot AI}},
  title        = {{Kimi K3}: Open Frontier Intelligence},
  year         = {2026},
  month        = jul,
  url          = {https://www.kimi.com/blog/kimi-k3}
}

@misc{mimov25,
  title={MiMo-V2.5},
  year={2026},
  howpublished={\url{https://huggingface.co/collections/XiaomiMiMo/mimo-v25}},
}

@misc{claudesonnet5,
  author       = {{Anthropic}},
  title        = {Introducing {Claude Sonnet 5}},
  year         = {2026},
  month        = jun,
  url          = {https://www.anthropic.com/news/claude-sonnet-5}
}

@misc{minimaxm3,
  author       = {{MiniMax}},
  title        = {{MiniMax M3}: Frontier Coding, 1M Context, Native
                  Multimodality---All in One Model},
  year         = {2026},
  month        = jun,
  url          = {https://www.minimax.io/blog/minimax-m3}
}

@Misc{smolagents,
  title =        {`smolagents`: a smol library to build great agentic systems.},
  author =       {Aymeric Roucher and Albert Villanova del Moral and Thomas Wolf and Leandro von Werra and Erik Kaunismäki},
  howpublished = {\url{https://github.com/huggingface/smolagents}},
  year =         {2025}
}

@misc{codexcli,
  author       = {{OpenAI}},
  title        = {{Codex CLI}},
  year         = {2026},
  url          = {https://learn.chatgpt.com/docs/codex/cli},
  note         = {Accessed July 27, 2026}
}

@misc{claudecode,
  author       = {{Anthropic}},
  title        = {How {Claude Code} Works},
  year         = {2026},
  url          = {https://code.claude.com/docs/en/how-claude-code-works},
  note         = {Accessed July 27, 2026}
}

@misc{grokbuild,
  author       = {{SpaceXAI}},
  title        = {{Grok Build}},
  year         = {2026},
  url          = {https://docs.x.ai/build/overview},
  note         = {Accessed July 27, 2026}
}

@article{mlestar,
  author       = {Jaehyun Nam and
                  Jinsung Yoon and
                  Jiefeng Chen and
                  Jinwoo Shin and
                  Sercan {\"{O}}. Arik and
                  Tomas Pfister},
  title        = {{MLE-STAR:} Machine Learning Engineering Agent via Search and Targeted
                  Refinement},
  journal      = {CoRR},
  volume       = {abs/2506.15692},
  year         = {2025}
}

@misc{teable,
  author       = {{Teable}},
  title        = {{Teable AI}: Overview},
  year         = {2026},
  url          = {https://help.teable.ai/en/basic/ai/overview},
  note         = {Accessed July 27, 2026}
}

@misc{deepanalyze,
        title={DeepAnalyze: Agentic Large Language Models for Autonomous Data Science}, 
        author={Shaolei Zhang and Ju Fan and Meihao Fan and Guoliang Li and Xiaoyong Du},
        year={2025},
        eprint={2510.16872},
        archivePrefix={arXiv},
        primaryClass={cs.AI},
        url={https://arxiv.org/abs/2510.16872}, 
  }

@article{taiji,
  author       = {Chao Zhang and
                  Shaolei Zhang and
                  Quehuan Liu and
                  Sibei Chen and
                  Tong Li and
                  Ju Fan},
  title        = {{TAIJI:} MCP-based Multi-Modal Data Analytics on Data Lakes},
  journal      = {CoRR},
  volume       = {abs/2505.11270},
  year         = {2025}
}

@inproceedings{aop,
  author       = {Jiayi Wang and
                  Guoliang Li},
  title        = {{AOP:} Automated and Interactive {LLM} Pipeline Orchestration for
                  Answering Complex Queries},
  booktitle    = {{CIDR}},
  publisher    = {www.cidrdb.org},
  year         = {2025}
}

@article{agenticdata,
  author       = {Ji Sun and
                  Guoliang Li and
                  Peiyao Zhou and
                  Yihui Ma and
                  Jingzhe Xu and
                  Yuan Li},
  title        = {AgenticData: An Agentic Data Analytics System for Heterogeneous Data},
  journal      = {CoRR},
  volume       = {abs/2508.05002},
  year         = {2025}
}

@inproceedings{deepeye,
author = {Li, Boyan and Peng, Yiran and Xie, Yupeng and Lu, Sirong and Zhu, Yizhang and Mu, Xing and Liu, Xinyu and Luo, Yuyu},
title = {DeepEye: A Steerable Self-driving Data Agent System},
year = {2026},
isbn = {9798400724503},
publisher = {Association for Computing Machinery},
address = {New York, NY, USA},
url = {https://doi.org/10.1145/3788853.3801612},
doi = {10.1145/3788853.3801612},
booktitle = {Companion of the International Conference on Management of Data},
pages = {74–77},
numpages = {4},
location = {India},
series = {SIGMOD Companion '26}
}

@article{datamagic,
  author       = {Yupeng Xie and
                  Chen Ma and
                  Zhenyang Wang and
                  Liangwei Wang and
                  Jiayi Zhu and
                  Chuxuan Zeng and
                  Zhouan Shen and
                  Boyan Li and
                  Yuyu Luo},
  title        = {DataMagic: Transforming Tabular Data into Data Insight Video},
  journal      = {CoRR},
  volume       = {abs/2606.20388},
  year         = {2026}
}

@article{xie2024haichart,
  title={Haichart: Human and AI paired visualization system},
  author={Xie, Yupeng and Luo, Yuyu and Li, Guoliang and Tang, Nan},
  journal={arXiv preprint arXiv:2406.11033},
  year={2024}
}

@article{xie2025visjudge,
  title={Visjudge-bench: Aesthetics and quality assessment of visualizations},
  author={Xie, Yupeng and Zhang, Zhiyang and Wu, Yifan and Lu, Sirong and Zhang, Jiayi and Yu, Zhaoyang and Wang, Jinlin and Hong, Sirui and Liu, Bang and Wu, Chenglin and others},
  journal={arXiv preprint arXiv:2510.22373},
  year={2025}
}

@article{tang2026igenbench,
  title={IGenBench: Benchmarking the Reliability of Text-to-Infographic Generation},
  author={Tang, Yinghao and Liu, Xueding and Zhang, Boyuan and Lan, Tingfeng and Xie, Yupeng and Lao, Jiale and Wang, Yiyao and Li, Haoxuan and Gao, Tingting and Pan, Bo and others},
  journal={arXiv preprint arXiv:2601.04498},
  year={2026}
}

@inproceedings{chen2025chartmark,
  title={ChartMark: A Structured Grammar for Chart Annotation},
  author={Chen, Yiyu and Wu, Yifan and Shen, Shuyu and Xie, Yupeng and Shen, Leixian and Xiong, Hui and Luo, Yuyu},
  booktitle={2025 IEEE Visualization and Visual Analytics (VIS)},
  pages={311--315},
  year={2025},
  organization={IEEE}
}

@article{tang2026vividoc,
  title={ViviDoc: Generating Interactive Documents through Human-Agent Collaboration},
  author={Tang, Yinghao and Xie, Yupeng and Feng, Yingchaojie and Lan, Tingfeng and Lao, Jiale and Cheng, Yue and Chen, Wei},
  journal={arXiv preprint arXiv:2603.27991},
  year={2026}
}

@article{bian2025you,
  title={You Don't Know Until You Click: Automated GUI Testing for Production-Ready Software Evaluation},
  author={Bian, Yutong and Lin, Xianhao and Xie, Yupeng and Liu, Tianyang and Zhuge, Mingchen and Lu, Siyuan and Tang, Haoming and Wang, Jinlin and Zhang, Jiayi and Chen, Jiaqi and others},
  journal={arXiv preprint arXiv:2508.14104},
  year={2025}
}

\appendix
\section{Additional Benchmark Details}
\label{app:additional-details}

\subsection{Cross-Language Transformation Details}
\label{app:cross-language-details}

This appendix expands the cross-language transformation in
Section~\ref{sec:cross-language}. We describe the construction contract,
algorithmic steps, intermediate artifacts, and normalized prompt templates.
Provider-specific system wrappers and batching delimiters are omitted because
they do not change the semantic contract. All generative calls use
deterministic decoding, and their outputs must pass the stated schema and
consistency checks before they can modify a database or workload.

\lstdefinestyle{appendixcode}{
    basicstyle=\ttfamily\scriptsize,
    columns=fullflexible,
    breaklines=true,
    breakatwhitespace=true,
    keepspaces=true,
    showstringspaces=false,
    frame=single,
    framerule=0.3pt,
    linewidth=\linewidth,
    xleftmargin=0pt,
    xrightmargin=0pt,
    framesep=3pt,
    aboveskip=0.5em,
    belowskip=0.6em
}

\subsubsection{Transformation Contract and Invariants}

The input is an executable Text-to-SQL tuple
\(T_0=(\database_0,q_0,\sigma_0)\), together with independently selected
database and question languages \((\ell_D,\ell_q)\). The output is
\(T_c=(\database_c,q_c,\sigma_c)\) plus a materialized replacement map \(M\).
The transformation enforces four invariants:

\begin{enumerate}[leftmargin=*]
    \item \textbf{Referential consistency.} Repeated representations of the
    same entity receive the same translation across primary keys, foreign
    keys, denormalized columns, and undeclared join paths.
    \item \textbf{Workload consistency.} Every schema identifier and value
    literal used by \(\sigma_c\) resolves against \(\database_c\).
    \item \textbf{Execution alignment.} Executing \(\sigma_c\) on
    \(\database_c\) reproduces the translated counterpart of the source
    result, including arity and row multiplicity.
    \item \textbf{Question fidelity.} The transformed question preserves the
    requested entities, predicates, aggregation, ordering, cardinality, units,
    and temporal scope of \(q_0\).
\end{enumerate}

The map \(M\), rather than an unconstrained model response, is the interface
between language generation and physical rewriting. Consequently, model calls
can be rerun or manually repaired without changing the deterministic migration
logic.

\subsubsection{Translation-Unit Extraction}

Let each text-bearing column be a vertex. We first add an explicit edge between
two columns when they participate in a declared foreign-key relationship or
share the same normalized name. Connected components of these edges form the
initial clusters \(\mathcal{C}_{\mathrm{exp}}\). To recover undeclared joins,
we collect the distinct non-null values \(V(C)\) of every cluster and compute
the overlap coefficient
\begin{equation}
    \rho(C_x,C_y)=
    \frac{|V(C_x)\cap V(C_y)|}
         {\min(|V(C_x)|,|V(C_y)|)}.
    \label{eq:cluster-overlap}
\end{equation}
Clusters are merged when \(\rho(C_x,C_y)>\theta_{\mathrm{ov}}\) and the
intersection contains at least \(k_{\min}\) distinct values. The second
condition prevents a single common token from connecting otherwise unrelated
columns. Both parameters are exposed in the construction configuration.

We then remove values that should not be localized. The filters cover nulls,
pure numbers, dates and timestamps, URLs, e-mail addresses, file paths,
machine-generated identifiers, standardized codes, and values already written
in the target language. Domain terms that must remain invariant can be added
to a protected glossary. Remaining values are deduplicated within a cluster
and divided into bounded prompt chunks. Every item retains its cluster ID and
all \((\text{table},\text{column})\) occurrences, so a single translated value
can later be fanned out consistently.

\paragraph{Extraction algorithm.}
The algorithm below separates clustering from prompt-size management; changing
the chunk budget \(B\) therefore does not change entity equivalence classes.

\begin{algorithm}[t]
\caption{Translation-unit extraction}
\label{alg:translation-unit-extraction}
\small
\begin{algorithmic}[1]
\Require Schema \(S\), database \(D\), target language \(\ell_D\)
\Require Thresholds \(\theta_{\mathrm{ov}},k_{\min}\)
\Require Prompt budget \(B\), protected glossary \(G\)
\Ensure Clusters \(\mathcal{C}\), schema units \(U_s\), value chunks \(U_v\)
\State \(V_T\gets\Call{TextColumns}{S}\)
\State \(\mathcal{C}\gets\Call{Singletons}{V_T}\)
\State \(\Call{UnionFKEndpoints}{\mathcal{C},S}\)
\State \(\Call{UnionSameNameColumns}{\mathcal{C},S}\)
\Repeat
    \State \(changed\gets\mathrm{false}\)
    \ForAll{unordered \((C_x,C_y)\in\mathcal{C}\)}
        \State \(I\gets V(C_x)\cap V(C_y)\)
        \State \(\rho\gets |I|/\min(|V(C_x)|,|V(C_y)|)\)
        \If{\(\rho>\theta_{\mathrm{ov}}\land |I|\ge k_{\min}\)}
            \State \(\Call{Union}{C_x,C_y}\); \(changed\gets\mathrm{true}\)
        \EndIf
    \EndFor
\Until{\(changed=\mathrm{false}\)}
\State \(U_s\gets\Call{ExtractSchemaUnits}{S,G}\)
\ForAll{\(C_j\in\mathcal{C}\)}
    \State \(V_j\gets\Call{DistinctNonNullValues}{D,C_j}\)
    \State \(V_j\gets\Call{FilterProtected}{V_j,\ell_D,G}\)
    \State \(V_j\gets\Call{DedupWithProvenance}{V_j}\)
    \State \(U_v\gets U_v\cup\Call{Chunk}{V_j,B}\)
\EndFor
\State \Return \(\mathcal{C},U_s,U_v\)
\end{algorithmic}
\end{algorithm}

\subsubsection{Replacement-Map Generation}

The materialized map contains three scoped mappings:
\begin{equation}
    M=\{M_{\mathrm{tab}},M_{\mathrm{col}},M_{\mathrm{val}}\},
\end{equation}
where table keys are database-scoped, column keys are table-scoped, and value
keys retain their column-cluster provenance. Figure~\ref{fig:replacement-map}
shows a normalized serialized form.

\begin{figure}[t]
\centering
\begin{minipage}{\columnwidth}
\begin{lstlisting}[style=appendixcode]
{
  "tables": [
    {"source": "...", "target": "..."}
  ],
  "columns": [
    {"table_source": "...", "source": "...", "target": "..."}
  ],
  "values": [
    {
      "cluster_id": "...",
      "source": "...",
      "target": "...",
      "occurrences": [
        {"table_source": "...", "column_source": "..."}
      ]
    }
  ]
}
\end{lstlisting}
\end{minipage}
\caption{Normalized serialized form of the replacement map.}
\label{fig:replacement-map}
\end{figure}

All responses are parsed as JSON and checked before map assembly. Each input
ID must occur exactly once, protected items must be unchanged, target table
names must be unique within a database, and target column names must be unique
within a table. A value shared by a cluster receives one target form, which is
copied to every listed occurrence. Invalid or incomplete responses are returned
to the model together with validator errors under a bounded retry policy; a
remaining conflict is repaired manually or the affected sample is rejected.
Figures~\ref{fig:prompt-schema-mapping}
and~\ref{fig:prompt-clustered-value} specify
the normalized contracts for the two mapping stages.

\begin{promptblock}{Prompt for schema mapping.}
{fig:prompt-schema-mapping}

\begin{lstlisting}[style=appendixcode]
SYSTEM
You localize relational schemas from <SOURCE_LANGUAGE> to
<TARGET_LANGUAGE>. Produce a faithful terminology map, not a new schema.

RULES
1. Translate only human-readable table and column names.
2. Preserve IDs, standardized codes, SQL keywords, and protected terms.
3. Preserve meaning, domain terminology, granularity, and abbreviations.
4. Table targets must be unique in the database. Column targets must be
   unique within their table.
5. Return JSON only. Do not add, remove, merge, or split input items.

INPUT
DOMAIN: <DATABASE_DOMAIN>
PROTECTED_GLOSSARY: <PROTECTED_GLOSSARY>
SCHEMA_ITEMS:
<LIST_OF_ITEMS_WITH_STABLE_IDS_TABLE_CONTEXT_AND_DESCRIPTIONS>

OUTPUT SCHEMA
{
  "items": [
    {"id": "<INPUT_ID>", "target": "<TRANSLATION>"}
  ]
}
\end{lstlisting}
\end{promptblock}

\begin{promptblock}{Prompt for clustered-value mapping.}
{fig:prompt-clustered-value}

\begin{lstlisting}[style=appendixcode]
SYSTEM
Translate database values from <SOURCE_LANGUAGE> to <TARGET_LANGUAGE>
while preserving equality and join semantics.

RULES
1. One source value has exactly one target value within this cluster.
2. Use the same target for every listed table/column occurrence.
3. Do not translate identifiers, codes, URLs, dates, numbers, or entries
   marked PROTECTED.
4. Preserve units, signs, precision, entity identity, and domain meaning.
5. Return JSON only and include every input ID exactly once.

CONTEXT
DOMAIN: <DATABASE_DOMAIN>
CLUSTER_ID: <CLUSTER_ID>
COLUMNS: <TABLE_COLUMN_OCCURRENCES>
SCHEMA_GLOSSARY: <RELEVANT_SCHEMA_MAP>
VALUES: <VALUES_WITH_STABLE_IDS_AND_PROTECTION_FLAGS>

OUTPUT SCHEMA
{
  "cluster_id": "<CLUSTER_ID>",
  "items": [
    {"id": "<INPUT_ID>", "target": "<TRANSLATION_OR_ORIGINAL>"}
  ]
}
\end{lstlisting}
\end{promptblock}

\subsubsection{Deterministic Database Migration}

The source database is copied before modification, and all changes are applied
inside a transaction. Rewrite order is important because map keys are expressed
in source-language identifiers. Cell values are updated first; columns are
renamed while source table names still exist; tables are renamed last. Dependent
views, indexes, triggers, foreign-key declarations, and schema metadata are
then rewritten against the target names. CHECK constraints that contain
translated enumerated values are temporarily removed and reconstructed with
their mapped literals. Any collision, unresolved reference, or constraint
failure aborts the transaction.

\begin{algorithm}[t]
\caption{Deterministic database migration}
\label{alg:database-migration}
\small
\begin{algorithmic}[1]
\Require Source database \(\database_0\), validated map \(M\)
\Ensure Translated database \(\database_c\), synchronized metadata \(S_c\)
\State \(\database_c\gets\Call{TransactionalCopy}{\database_0}\)
\State \(K\gets\Call{ExtractDefinitions}{\database_c}\)
\State \(\Call{DeferAffectedChecks}{\database_c,K,M}\)
\ForAll{\((t,c,v\mapsto v')\in M_{\mathrm{val}}\)}
    \State \(\Call{UpdateExactValue}{\database_c,t,c,v,v'}\)
\EndFor
\ForAll{\((t,c\mapsto c')\in M_{\mathrm{col}}\)}
    \State \(\Call{RenameColumn}{\database_c,t,c,c'}\)
\EndFor
\ForAll{\((t\mapsto t')\in M_{\mathrm{tab}}\)}
    \State \(\Call{RenameTable}{\database_c,t,t'}\)
\EndFor
\State \(K_c\gets\Call{RewriteDefinitions}{K,M}\)
\State \(\Call{RestoreDefinitions}{\database_c,K_c}\)
\State \(S_c\gets\Call{SyncMetadata}{\database_c,M}\)
\State \(\Call{AssertNoDrift}{\database_c,S_c,M}\)
\State \(\Call{Commit}{\database_c}\)
\State \Return \(\database_c,S_c\)
\end{algorithmic}
\end{algorithm}

Updates are scoped by the source table and column recorded in \(M\); global
string replacement is never applied to database contents. When a source value
is a substring of another value, exact matching is used for cells and
longest-first matching is used only inside parsed schema definitions.

\subsubsection{Protect--Replace--Restore SQL Rewriting}

SQL rewriting is a workload-migration operation over \(\sigma_0\), not a new
query-generation call. We parse nested query blocks and build a scope table for
base tables, CTEs, aliases, and projected columns. String literals, CTE names,
alias declarations, and table-name positions are replaced with typed markers.
Schema identifiers are then mapped according to their resolved scope. A string
literal is translated only when its comparison context resolves to a column
cluster containing the corresponding value entry. This prevents a surface form
that appears in two unrelated columns from receiving the wrong replacement.
Subqueries are rewritten from the innermost scope outward, after which markers
are restored and the resulting SQL is parsed again.

\begin{algorithm}[t]
\caption{Protect--replace--restore SQL rewriting}
\label{alg:sql-rewriting}
\small
\begin{algorithmic}[1]
\Require SQL \(\sigma_0\), replacement map \(M\)
\Require Source schema \(S_0\), target schema \(S_c\)
\Ensure Rewritten SQL \(\sigma_c\)
\State \((ast,scopes)\gets\Call{ParseResolve}{\sigma_0,S_0}\)
\State \((protected,markers)\gets\Call{Protect}{ast}\)
\State \(ordered\gets\Call{InnermostToOutermost}{scopes}\)
\ForAll{\(scope\in ordered\)}
    \State \(\Call{ReplaceTables}{scope,M_{\mathrm{tab}}}\)
    \State \(\Call{ReplaceColumns}{scope,M_{\mathrm{col}}}\)
    \ForAll{literal \(L\) with resolved context \((t,c)\)}
        \If{\((t,c,L)\in M_{\mathrm{val}}\)}
            \State \(markers[L]\gets M_{\mathrm{val}}[(t,c,L)]\)
        \EndIf
    \EndFor
\EndFor
\State \(\sigma_c\gets\Call{RestoreSerialize}{protected,markers}\)
\State \(\Call{AssertParseable}{\sigma_c}\)
\State \(\Call{AssertResolved}{\sigma_c,S_c}\)
\State \Return \(\sigma_c\)
\end{algorithmic}
\end{algorithm}

The protection layer also prevents accidental substitutions inside SQL
keywords, function names, numeric constants, and partial identifiers. When an
unqualified column is ambiguous under the current scope, the rewrite is not
guessed; it is marked for repair.

\subsubsection{Question Translation and Language Composition}

Question language is varied independently of database language. The question
translator receives the original question, a focused glossary containing only
entities relevant to that sample, and optional SQL alignment context. The SQL
is construction-time context: it constrains preservation of operators and
conditions but is not copied into the natural-language output. No database or
SQL artifact is modified during this step. Verified question variants are
joined to verified database/SQL variants by stable sample ID, which allows
\((\ell_q,\ell_D)\) combinations to be assembled without repeating database
migration.
The normalized translation contract is given in
Figure~\ref{fig:prompt-question-translation}.

\begin{promptblock}{Prompt for question translation.}
{fig:prompt-question-translation}

\begin{lstlisting}[style=appendixcode]
SYSTEM
Translate an analytical question from <SOURCE_LANGUAGE> to
<TARGET_LANGUAGE>. Preserve the exact answer semantics.

PRESERVE
- entities, value literals, units, signs, and numerical thresholds;
- filters, negation, conjunction, and comparison direction;
- aggregation, grouping, distinctness, ranking, ordering, and limits;
- time windows, inclusivity of boundaries, and requested output fields.

RULES
1. Use the supplied target-language glossary for schema and value terms.
2. Write a natural user question; do not mention SQL, schemas, or this task.
3. Do not add explanations, assumptions, or answer values.
4. Return JSON only.

INPUT
SOURCE_QUESTION: <QUESTION>
FOCUSED_GLOSSARY: <RELEVANT_TABLE_COLUMN_VALUE_MAP>
OPTIONAL_SQL_ALIGNMENT_CONTEXT: <GOLD_SQL_OR_OPERATOR_SIGNATURE>

OUTPUT SCHEMA
{
  "question": "<TRANSLATED_QUESTION>",
  "used_mapping_ids": ["<ID>"]
}
\end{lstlisting}
\end{promptblock}

\subsubsection{Validation Gates and Failure Handling}

Validation proceeds from inexpensive structural checks to semantic checks:

\begin{enumerate}[leftmargin=*]
    \item \textbf{Map validation} checks coverage, scope, protected identities,
    and table/column target-name collisions.
    \item \textbf{Database validation} checks that mapped entities exist,
    dependent definitions resolve, and integrity constraints can be applied.
    \item \textbf{Workload validation} parses \(\sigma_c\), resolves every
    identifier, executes it on \(\database_c\), and compares the result with
    \(\tau_M(\operatorname{Exec}(\database_0,\sigma_0))\). Comparison preserves
    duplicate multiplicity; row order is enforced when specified by the query.
    \item \textbf{Question validation} uses an LLM judge to check that \(q_c\)
    remains answerable by \(\sigma_c\) and preserves its analytical intent.
\end{enumerate}
Figure~\ref{fig:prompt-question-alignment} instantiates the final semantic
audit.

\begin{promptblock}{Prompt for question--SQL alignment.}
{fig:prompt-question-alignment}

\begin{lstlisting}[style=appendixcode]
SYSTEM
Audit whether a translated analytical question preserves the source intent
and remains answered by the translated SQL. Do not solve the query.

CHECK
1. Requested output entities and fields.
2. All filters, values, comparison directions, and negations.
3. Aggregation, grouping, DISTINCT semantics, ordering, ranking, and limits.
4. Time windows, boundary inclusivity, units, and numerical scale.
5. Consistency with the supplied schema/value glossary.

INPUT
SOURCE_QUESTION: <SOURCE_QUESTION>
TRANSLATED_QUESTION: <TRANSLATED_QUESTION>
SOURCE_SQL: <SOURCE_SQL>
TRANSLATED_SQL: <TRANSLATED_SQL>
FOCUSED_GLOSSARY: <RELEVANT_MAP>

OUTPUT SCHEMA
{
  "status": "pass | repair | reject",
  "issues": [
    {"type": "<ISSUE_TYPE>", "description": "<DESCRIPTION>"}
  ],
  "revised_question": "<ONLY_IF_REPAIR>"
}
\end{lstlisting}
\end{promptblock}

A tuple advances only when all gates pass. A failed map is regenerated or
edited before any rewrite. A failed SQL is repaired against the same map and
re-executed. A repaired question is subjected to the alignment prompt again.
Cases that cannot be made execution-aligned and semantically faithful are
discarded. The accepted stage artifact contains \((\database_c,q_c,\sigma_c)\),
the replacement map, synchronized metadata, and validation status; the final
question--workspace answerability check is performed later by the expert
review process in Section~\ref{sec:human-review}.

\subsection{Constraint-Aware Relational Sampling Details}
\label{app:sampling-details}

This appendix expands the relational sampling stage in
Section~\ref{sec:data-sampling}. The procedure constructs a new task-local
database instance rather than approximating the answer obtained from the
source database. Accordingly, the source and sampled answers may differ in
their entities, multiplicities, aggregate values, and ordering. Executability
and task validity are checked on the sampled instance itself.

\subsubsection{Sampling Contract}

The input is the verified cross-language tuple
\(T_c=(\database_c,q_c,\sigma_c)\). Let \(\mathcal{T}_c\) be the complete table
inventory of \(\database_c\), and let \(\database_c[R]\) denote the rows of
table \(R\). A sampling policy
\begin{equation}
    \psi=\bigl(z,\{b_R:R\in\mathcal{T}_c\},K,R_a,\mathcal{E}^{+}\bigr)
    \label{eq:sampling-policy}
\end{equation}
contains a base random seed \(z\), a soft row budget \(b_R\) for each table, a
maximum number of attempts \(K\), an optional anchor table \(R_a\), and schema
relationships \(\mathcal{E}^{+}\) not declared in the source database. The
latter covers, for example, known key pairs in databases with incomplete
foreign-key metadata. Budgets bound ordinary random additions; protected rows
and rows introduced by relational closure take precedence and may exceed them.
We use \(K=3\) for the benchmark construction.

Every sampled instance preserves the complete table and column inventory:
\begin{equation}
    \operatorname{Tables}(\database_s)=\mathcal{T}_c,\qquad
    \operatorname{Schema}(\database_s[R])
    =\operatorname{Schema}(\database_c[R]).
    \label{eq:sampling-schema-invariant}
\end{equation}
Thus, sampling changes table contents but does not use the source SQL to
remove tables or columns. Empty source tables remain valid empty tables with
their schema intact.

\subsubsection{Safeguard Extraction}

We form the safeguard set
\(\mathcal{C}_s=\mathcal{C}_{\mathrm{schema}}\cup
\mathcal{C}_{\mathrm{query}}\). The schema component is represented as a
directed relationship graph \(G_s=(\mathcal{T}_c,\mathcal{E}_s)\). Each edge
records a child table and column, the corresponding parent table and column,
and whether the relationship is declared or supplied by
\(\mathcal{E}^{+}\). Composite keys are retained as tuples rather than
decomposed into independent column constraints.

The query component is extracted from a resolved SQL AST. Name resolution is
performed separately inside each query block so that aliases, common-table
expressions, and correlated subqueries do not create spurious bindings. We
record four types of safeguard:

\begin{enumerate}[leftmargin=*]
    \item \textbf{Predicate bindings} associate a resolved column with a
    literal or literal set used by equality, membership, range, or pattern
    predicates. Literal types are preserved during matching.
    \item \textbf{Query relationships} record the resolved column pairs in
    explicit and implicit joins, including multi-column join keys.
    \item \textbf{Boundary bindings} retain values that define temporal or
    numerical intervals.
    \item \textbf{Target bindings} identify explicitly named entities whose
    disappearance would change the referent of the question.
\end{enumerate}

Unsupported expressions do not trigger string-based guessing. They are
retained in the AST record and left to execution validation; samples for which
the protected values or join endpoints cannot be resolved are marked for
repair before sampling.

\subsubsection{Anchor Rows and Soft Budgets}

For each table \(R\), the sampler first constructs an anchor set \(A_R\).
Rows matching protected equality, membership, or target bindings are inserted
directly. For range and pattern predicates, matching rows are selected using
the same typed operator as the SQL expression. When a predicate is attached
to a nested query, its anchors remain scoped to the base relation resolved in
that query block. When \(R_a\) is configured, its selected rows provide the
starting keys for relationship propagation. Anchor selection therefore
operates on database values, not surface-form occurrences in serialized rows.

The remaining capacity is filled by sampling within each table. For a
table \(R\) whose anchors do not exhaust its budget \(b_R\), the initial row set
is
\begin{equation}
    S_R^{(0)}=A_R\cup
    \operatorname{Sample}(\database_c[R]\setminus A_R,\,b_R-|A_R|).
    \label{eq:initial-table-sample}
\end{equation}
Here, \(\operatorname{Sample}(X,n)\) returns up to \(n\) rows from \(X\), so
small tables are retained in full. If the anchors already meet or exceed the
budget, no additional rows are sampled. Tables not referenced by \(\sigma_c\)
use the same rule with an empty anchor set.

\subsubsection{Relational Closure}

Independent table samples can contain a selected foreign key without its
referenced row or leave a protected query join without any matched pair. We
therefore augment the initial row sets by relational closure. For every
selected child row, its non-null referenced key introduces the matching parent
row. For query relationships, the sampler additionally retains matched rows
along the protected join path from the current anchors. Newly introduced rows
are placed back on the closure queue until no relationship adds a row. Since
closure only adds rows drawn from the finite source instance, the procedure
terminates even when the schema graph contains cycles.

One-to-many expansion is bounded for ordinary random rows. Anchor-derived and
query-path matches have priority; optional matches are drawn with the
relationship-specific random stream until the relevant soft budget is
reached. Missing referenced keys, type-incompatible join columns, and an empty
protected join are recorded as validation errors rather than repaired by
fabricating records.

Algorithm~\ref{alg:relational-sampling} summarizes the complete procedure.

\begin{algorithm}[t]
\caption{Constraint-aware relational sampling}
\label{alg:relational-sampling}
\small
\begin{algorithmic}[1]
\Require Verified tuple \((\database_c,q_c,\sigma_c)\), policy \(\psi\)
\Ensure Sampled database \(\database_s\), reference result \(\goldrel_s\), ledger \(L_s\)
\State \(\mathcal{C}_{\mathrm{schema}}\gets\Call{SchemaRelations}{\database_c,\mathcal{E}^{+}}\)
\State \(\mathcal{C}_{\mathrm{query}}\gets\Call{QuerySafeguards}{\sigma_c,\database_c}\)
\State \(\mathcal{C}_s\gets\mathcal{C}_{\mathrm{schema}}\cup\mathcal{C}_{\mathrm{query}}\)
\For{attempt \(k=1,\ldots,K\)}
    \State \(A\gets\Call{AnchorRows}{\database_c,\mathcal{C}_{\mathrm{query}},R_a,z,k}\)
    \State \(S\gets\Call{SeededTableSamples}{\database_c,A,\psi,k}\)
    \State \(S\gets\Call{RelationalClosure}{S,\mathcal{C}_s}\)
    \State \(\database_s\gets\Call{MaterializeAllSchemas}{\database_c,S}\)
    \State \(v_s\gets\Call{ValidateStructure}{\database_s,\mathcal{C}_{\mathrm{schema}}}\)
    \If{\(v_s=\mathrm{pass}\)}
        \State \((e_s,\goldrel_s)\gets\Call{Execute}{\database_s,\sigma_c}\)
        \State \(v_q\gets\Call{ValidateTask}{e_s,\goldrel_s,\sigma_c,\mathcal{C}_{\mathrm{query}}}\)
        \If{\(v_q=\mathrm{pass}\)}
            \State \(L_s\gets\Call{BuildLedger}{\database_s,\goldrel_s,\psi,\mathcal{C}_s,k}\)
            \State \Return \(\database_s,\goldrel_s,L_s\)
        \EndIf
    \EndIf
\EndFor
\State \Return \(\mathrm{reject}\)
\end{algorithmic}
\end{algorithm}

\subsubsection{Materialization and Validation}

Materialization creates a fresh SQLite database, recreates every source table
with its column types and key declarations, inserts the selected rows, and
rebuilds applicable indexes and views. Inserts occur inside a transaction with
foreign-key checking enabled at validation time. The resulting database is
then checked in two stages.

\paragraph{Structural validation.}
We verify table and column inventory, declared primary-key uniqueness,
foreign-key consistency, row serializability, and the presence of protected
bindings and query-path matches. We also compare per-table row counts against
the recorded selected-row sets; this catches silent insertion loss caused by
type conversion or duplicate handling.

\paragraph{Execution validation.}
The transformed SQL \(\sigma_c\) is parsed and executed on \(\database_s\).
Successful execution defines the candidate answer
\(\goldrel_s=\operatorname{Exec}(\database_s,\sigma_c)\); no equality test
against the source answer is applied. We check the returned arity against the
resolved projection and record query-signature diagnostics for conditions such
as a broken join, a null-only aggregate, or fewer candidates than a requested
ranking. These conditions are not universal rejection rules: an empty or
null-valued answer can be semantically correct. An unexpected empty result
caused by the loss of protected predicates or join matches fails the attempt;
ambiguous cases are forwarded to the expert review stage.

\subsubsection{Retry, Repair, and Provenance}

A failed attempt is retried with a deterministically derived seed. Failures
caused by insufficient optional coverage are handled by resampling; failures
caused by unresolved relationships, missing protected values, or unsuitable
budgets require configuration repair before another attempt. A task is
discarded when it exhausts the configured attempt budget or cannot produce an
executable and meaningful relational instance.

For every accepted task, the internal sampling ledger stores the source task
and database identifiers, SQL hash, policy and attempt seed, per-table soft
budgets, extracted safeguards, added relationship pairs, row counts before and
after sampling, retry history, intermediate-database hash, execution status,
and candidate-answer hash. The ledger is used to reproduce construction and
audit later repairs; it is not included in the agent-visible workspace.

\subsection{Modality Routing \& Artifact Rendering Details}
\label{app:artifact-rendering-details}

This appendix expands Section~\ref{sec:artifact-rendering}. We first specify
query-independent base routing and verify each rendered relation against the
task-local intermediate database. We then detail fact-grounded document
rendering and the separate query-conditioned video branch, and trace both
procedures with released tasks. The algorithms describe the construction
interfaces; renderer-specific templates and visual themes can change without
modifying their grounding and validation contracts.

\subsubsection{Query-Independent Base Routing and Materialization}

\paragraph{Base-renderer assignment.}
For every sampled table \(R\in\database_s\), the base router extracts a
descriptor \(m_R\) containing its schema, column types, key declarations, row
and column counts, missing-value profile, and serializability constraints. It
does not inspect the question, task SQL, or candidate answer. Its renderer set
is

\begin{equation}
    \mathcal{F}_{\mathrm{base}}
    =\{\mathtt{CSV},\mathtt{JSON},\mathtt{SQLite},
    \mathtt{Markdown},\mathtt{PDF}\},
\end{equation}

Video is not a member of \(\mathcal{F}_{\mathrm{base}}\): it operates at the
task level and may depend on the question, SQL structure, and candidate answer
rather than materializing a single table \(R\). The base policy first removes
renderers that cannot faithfully encode \(m_R\).
Flat tables can be written as CSV or record-oriented JSON; relational groups
with declared keys can be retained in SQLite; and tables selected for
long-form rendering are passed to the Markdown/PDF document generator. A
fixed seed breaks ties among eligible choices, while batch-level counters
favor underrepresented modalities. These counters balance assignments across
the collection; they do not prescribe the number of files, artifact sizes, or
the proportions of long and wide tables in an individual workspace.

Algorithm~\ref{alg:artifact-routing} gives the normalized procedure. A table
may be assigned to more than one renderer, and every table must have at least
one successful base representation. When a preferred renderer fails its
round-trip check, the router tries the next compatible choice; failure of all
choices rejects the materialization attempt.

\begin{algorithm}[t]
\caption{Query-independent base routing and materialization}
\label{alg:artifact-routing}
\small
\begin{algorithmic}[1]
\Require Sampled database \(\database_s\), policy \(\pi_r\), seed \(z_r\)
\Require Base renderer set \(\mathcal{F}_{\mathrm{base}}\), coverage state \(H\)
\Require Fallback order \(O_f\)
\Ensure Base workspace \(\workspace_{\mathrm{base}}\)
\State \(\workspace_{\mathrm{base}}\gets\varnothing\)
\ForAll{\(R\in\Call{SeededOrder}{\database_s,z_r}\)}
    \State \(m_R\gets\Call{TableMetadata}{R}\)
    \State \(E_R\gets\Call{Compatible}{m_R,\mathcal{F}_{\mathrm{base}}}\)
    \State \(U_R\gets\Call{Route}{m_R,E_R,H,\pi_r,z_r}\)
    \If{\(U_R=\varnothing\)}
        \State \(U_R\gets\{\Call{First}{E_R}\}\)
    \EndIf
    \ForAll{\(f\in\Call{WithFallbacks}{U_R,E_R,O_f}\)}
        \State \(a\gets\Call{Render}{R,f}\)
        \If{\(\Call{RoundTripValid}{a,R,f}\)}
            \State \(\workspace_{\mathrm{base}}\gets
                \workspace_{\mathrm{base}}\cup\{a\}\)
            \State \(H[f]\gets H[f]+1\)
            \State \(\Call{MarkRepresented}{R}\)
            \If{\(\Call{AssignmentsComplete}{R,U_R}\)}
                \State \textbf{break}
            \EndIf
        \EndIf
    \EndFor
    \If{\(\neg\Call{Represented}{R}\)}
        \State \Return \(\mathrm{reject}\)
    \EndIf
\EndFor
\State \Return \(\workspace_{\mathrm{base}}\)
\end{algorithmic}
\end{algorithm}

\paragraph{Serialization contract.}
Each structured renderer is paired with a parser that reconstructs a
canonical relation. The comparison covers the column inventory, typed cell
values, null positions, duplicate-row multiplicity, and row count. CSV uses
quoted fields and an explicit schema side record during construction; JSON is
written as a list of records with stable keys; and SQLite recreates declared
types, primary keys, and foreign keys before inserting rows. File order is not
treated as relational meaning unless an order-bearing field is itself part of
the table. For a renderer \(f\), acceptance requires

\begin{equation}
    \operatorname{Canon}
    \bigl(\operatorname{Parse}_f(\operatorname{Render}_f(R))\bigr)
    =\operatorname{Canon}(R),
    \label{eq:artifact-roundtrip}
\end{equation}

where canonicalization normalizes physical encodings while retaining data
types and row multiplicity. The construction ledger records the table,
renderer, seed, file hash, row count, and validation result. This information
supports reproduction and repair but is not placed in the released
workspace.

\subsubsection{Fact-Grounded Table-to-Document Rendering}

\paragraph{Document plan.}
Document rendering converts a complete sampled table into a long-form report
without placing the full table in a prompt at once. The LLM-based planner
first infers a domain-appropriate genre, such as a clinical event ledger,
financial audit, or operational briefing. It then chooses anchor columns that
identify the entity or record. Declared keys receive priority; otherwise,
high-uniqueness identifiers and stable entity names are used. The remaining
columns are grouped by semantic role, and rows are partitioned to respect the
generation budget. A small table is represented by a single row batch and
attribute cluster, making one-pass generation a special case of the same
procedure.
For larger tables, the plan can emphasize rows (a horizontal pass), attribute
clusters (a vertical pass), or alternate between them. In each case, the
generation unit reduces to anchors, a bounded row set, and one or more
attribute clusters.

Every generation block contains (i) its row identifiers, (ii) the anchor
columns repeated from those rows, and (iii) one attribute cluster. Repeating
anchors allows sections generated from distant attribute clusters to remain
joinable after assembly. Each non-null source cell is assigned an internal
cell ID, and each block carries the set of cell IDs that it must express.
Null-valued fields follow the plan's explicit policy: either state that the
field is unavailable or omit it without inventing a replacement.

\paragraph{Grounded generation.}
Algorithm~\ref{alg:document-rendering} shows the generation and checking loop.
The model may vary discourse structure, connective text, and non-evidential
background detail, but protected identifiers, numbers, dates, units, and
categorical values must remain recoverable with their source precision. When
contrastive or corrective prose is used for stress testing, the source value
must be identified unambiguously as the final record value. A failed block is
regenerated from the validator feedback rather than silently removed.

\paragraph{Controlled document complexity.}
Three controls vary extraction difficulty without changing the sampled table.
Recursive detailing expands a block into successively more specific report
sections. The null policy varies whether missing fields are stated or omitted.
Finally, a configurable subset of blocks receives domain-relevant narrative
context, nearby non-answer facts, or a correction-style presentation. Such
additions cannot replace a required source cell, alter a protected value, or
introduce a competing final value for the same record and field.

\begin{algorithm}[t]
\caption{Fact-grounded table-to-document rendering}
\label{alg:document-rendering}
\small
\begin{algorithmic}[1]
\Require Sampled table \(X_R\), schema metadata \(m_R\)
\Require Token budget \(B\), retry limit \(K\), output format \(f\)
\Ensure Markdown or PDF artifact \(d_R\)
\State \((\eta_R,\mathcal{K}_R,\mathcal{C}_R,\mathcal{I}_R)
    \gets\Call{PlanDocument}{X_R,m_R,B}\)
\State \(\mathcal{B}_R\gets
    \Call{BuildBlocks}{X_R,\mathcal{K}_R,\mathcal{C}_R,\mathcal{I}_R}\)
\State \(\mathcal{G}_R\gets\varnothing\)
\ForAll{\(B_j\in\mathcal{B}_R\)}
    \For{\(k=1\) to \(K\)}
        \State \(g_j\gets\Call{GenerateSection}{B_j,\eta_R}\)
        \State \(e_j\gets\Call{ValidateCells}{g_j,B_j}\)
        \If{\(e_j=\varnothing\)}
            \State \textbf{break}
        \EndIf
        \State \(B_j\gets\Call{AttachFeedback}{B_j,e_j}\)
    \EndFor
    \If{\(e_j\ne\varnothing\)}
        \State \Return \(\mathrm{reject}\)
    \EndIf
    \State \(\mathcal{G}_R\gets\mathcal{G}_R\cup\{g_j\}\)
\EndFor
\State \(d_R\gets\Call{AssembleMarkdown}{\mathcal{G}_R,\eta_R}\)
\State \(\Call{ValidateCoverage}{d_R,X_R}\)
\If{\(f=\mathtt{PDF}\)}
    \State \(d_R\gets\Call{ConvertAndCheckPdf}{d_R}\)
\EndIf
\State \Return \(d_R\)
\end{algorithmic}
\end{algorithm}

The block validator combines exact and typed checks. Exact-token checks cover
IDs, codes, protected strings, and categorical values. Numeric values are
parsed with their units and compared at the recorded precision; dates and
times are normalized before comparison; and ordinary text values are checked
against the source cell or an approved surface-form map. The final coverage
pass verifies that every required cell ID is linked to at least one document
span and that no span assigns two incompatible values to the same
record--field pair. PDF conversion is followed by text extraction and page
decoding checks. Figure~\ref{fig:prompt-document-generation} gives the
normalized block-level generation contract.

\begin{promptblock}{Prompt for fact-grounded document generation.}
{fig:prompt-document-generation}

\begin{lstlisting}[style=appendixcode]
SYSTEM
Write one section of a realistic <DOCUMENT_GENRE> from the supplied
table block. The table block is the sole source of record facts.

GROUNDING RULES
1. Express every REQUIRED_CELL_ID with its exact entity association.
2. Preserve identifiers, categorical values, numeric precision, dates,
   times, signs, and units. Do not merge values from different rows.
3. Follow NULL_POLICY. Never infer a missing value.
4. You may add connective prose or domain-neutral scene detail, but it
   must not create a competing value for any protected field.
5. Keep ANCHOR_FIELDS explicit so this section can be joined with other
   sections about the same records.
6. Return only the requested section and the cell-to-span alignment JSON.

INPUT
STYLE: <STYLE_SPECIFICATION>
ANCHOR_FIELDS: <ANCHOR_COLUMNS_AND_VALUES>
ATTRIBUTE_CLUSTER: <COLUMN_NAMES_TYPES_AND_DESCRIPTIONS>
ROWS: <TYPED_VALUES_WITH_CELL_IDS>
REQUIRED_CELL_IDS: <CELL_IDS>
NULL_POLICY: <EXPLICIT_OR_OMIT>

OUTPUT
{
  "section": "<MARKDOWN_SECTION>",
  "alignments": [
    {"cell_id": "<CELL_ID>", "surface": "<TEXT_SPAN>"}
  ]
}
\end{lstlisting}
\end{promptblock}

\subsubsection{Document-Rendering Running Example}
\label{app:document-running-example}

Figure~\ref{fig:document-running-example} traces one row from the
\texttt{LABEVENTS} document in released Task~310. The sampled row contains
seven fields. The planner uses \texttt{ROW\_ID} as the anchor and separates
event context from the laboratory measurement. The two resulting blocks are
therefore generated at different positions in a long clinical ledger but can
be joined through event 142456.

\begin{figure*}[t]
\centering
\small
\setlength{\tabcolsep}{4pt}
\begin{tabular}{@{}p{0.25\textwidth}
    @{\hspace{0.5em}\(\Longrightarrow\)\hspace{0.5em}}
    p{0.28\textwidth}
    @{\hspace{0.5em}\(\Longrightarrow\)\hspace{0.5em}}
    p{0.31\textwidth}@{}}
\toprule
\multicolumn{3}{c}{%
\textbf{Document plan:}
clinical event ledger \(\mid\)
\textcolor{docanchor}{anchor \(\mathcal{K}_R=\{\texttt{ROW\_ID}\}\)} \(\mid\)
\textcolor{doccontext}{context cluster \(C_1\)} \(\mid\)
\textcolor{docmeasure}{measurement cluster \(C_2\)}} \\
\midrule
\centering\textbf{Sampled row}\par\raggedright\arraybackslash
\begin{itemize}[leftmargin=1.2em,itemsep=0.25em,topsep=0.35em]
    \item \texttt{ROW\_ID}:
    \textcolor{docanchor}{\textbf{142456}}
    \item \texttt{SUBJECT\_ID}:
    \textcolor{doccontext}{\textbf{23070}}
    \item \texttt{HADM\_ID}:
    \textcolor{doccontext}{\textbf{127721}}
    \item \texttt{CHARTTIME}:
    \textcolor{doccontext}{\textbf{2105-12-09 20:58:00}}
    \item \texttt{ITEMID}:
    \textcolor{docmeasure}{\textbf{50954}}
    \item \texttt{VALUENUM}:
    \textcolor{docmeasure}{\textbf{151.0}}
    \item \texttt{VALUEUOM}:
    \textcolor{docmeasure}{\textbf{iu/l}}
\end{itemize}
&
\centering\textbf{Grounded blocks}\par\raggedright\arraybackslash
\begin{itemize}[leftmargin=1.2em,itemsep=0.55em,topsep=0.35em]
    \item \(B_1\): event
    \textcolor{docanchor}{\textbf{142456}}; patient
    \textcolor{doccontext}{\textbf{23070}}; admission
    \textcolor{doccontext}{\textbf{127721}}; timestamp
    \textcolor{doccontext}{\textbf{2105-12-09 20:58:00}}.
    \item \(B_2\): event
    \textcolor{docanchor}{\textbf{142456}}; item
    \textcolor{docmeasure}{\textbf{50954}}; result
    \textcolor{docmeasure}{\textbf{151.0 iu/l}}.
\end{itemize}
The anchor is repeated so that independently generated sections remain
joinable.
&
\centering\textbf{Released document spans}\par\raggedright\arraybackslash
\begin{itemize}[leftmargin=1.2em,itemsep=0.55em,topsep=0.35em]
    \item ``Regarding lab event
    \textcolor{docanchor}{\textbf{142456}}, this record pertains to patient
    \textcolor{doccontext}{\textbf{23070}} during hospital admission
    \textcolor{doccontext}{\textbf{127721}} \ldots{}
    \textcolor{doccontext}{\textbf{December 9th, 2105, at 20:58:00}}.''
    \item ``For event
    \textcolor{docanchor}{\textbf{142456}}, the test ordered under item code
    \textcolor{docmeasure}{\textbf{50954}} \ldots{} yielded a result of
    \textcolor{docmeasure}{\textbf{151.0}} \ldots{}
    \textcolor{docmeasure}{\textbf{iu/l}}.''
\end{itemize}
\\
\midrule
\multicolumn{3}{c}{%
\textcolor{docanchor}{\textbf{anchor / record identity}}
\(\quad\)
\textcolor{doccontext}{\textbf{event context}}
\(\quad\)
\textcolor{docmeasure}{\textbf{laboratory measurement}}} \\
\bottomrule
\end{tabular}
\caption{Source-to-document alignment for one sampled
\texttt{LABEVENTS} row in Task~310. Colors preserve field-group provenance
from the source row through grounded generation blocks to mentions in the
released Markdown document; ellipses shorten the displayed spans.}
\label{fig:document-running-example}
\end{figure*}

All seven source cells are aligned to the two spans, with the anchor appearing
in both. Recovering the measurement as a structured record requires
associating fields across separated narrative sections through the exact
record identifier while ignoring surrounding clinical and operational prose.

\subsubsection{Query-Conditioned Video Rendering}

\paragraph{Evidence selection.}
Video augmentation begins only after the sampled SQL has been executed. The
selector parses comparison predicates, projected fields, aggregation and
ordering operators, and the candidate tabular result. A predicate candidate
is represented as a typed atom

\begin{equation}
    e^{\mathrm{pred}}=
    (t,c,o,v,\lambda,\gamma),
\end{equation}

where \(t\) and \(c\) identify the source table and column, \(o\) is the
operator, \(v\) is the typed condition value, \(\lambda\) is its aligned
question span, and \(\gamma\) records display constraints such as precision
and units. Null tests, join keys, grouping operators, unstable relative-time
conditions, and projection-only fields are excluded from predicate
abstraction. An answer atom

\begin{equation}
    e^{\mathrm{ans}}=(r,c,v,\gamma)
\end{equation}

identifies one result row, output field, typed value, and display contract.
Answer-evidence rendering is used only when the result is compact enough for
its required atoms to remain legible across scenes.

\paragraph{Two rendering strategies.}
In \emph{predicate abstraction}, one or more stable predicate atoms are
removed from the explicit question and expressed through a business scene,
such as a configuration panel, time window, or eligibility rule. The remaining
workspace still contains the records on which the recovered predicate must be
applied. In \emph{answer-evidence rendering}, result atoms are distributed
across tables, charts, cards, or temporally separated views. Direct copies of
the same answer-bearing source are withheld when they would make the video
unnecessary. In both cases, distractors are drawn from nearby fields, entities,
periods, or boundary examples; they cannot change the selected atom or create
a second valid answer.

\paragraph{Storyboard and rendering.}
Each selected atom receives a stable evidence ID. A storyboard scene lists its
duration, evidence IDs, visual component, exact data constants, visible text,
and narration. The planner can decide how an atom is communicated, but the
constants in its component specification are filled from the typed evidence
record. Supporting records for boundary cases and distractors are queried from
the sampled database and retain their source table and row identifiers. The
storyboard is compiled into a task-specific React/Remotion composition.
Narration is synthesized separately, and measured audio duration is used to
set scene timing. Compilation or rendering failures are repaired at the
component level and rendered again in task isolation.

Algorithm~\ref{alg:video-rendering} summarizes the branch. Question rewriting
operates over the aligned span \(\lambda\): it replaces only the selected
condition or answer reference with a description that points to the video.
The requested output fields, aggregation, comparison, ordering, units, and
temporal scope remain unchanged. The question is rejected when the selected
span cannot be removed without changing those semantics.

\begin{algorithm}[t]
\caption{Query-conditioned video augmentation}
\label{alg:video-rendering}
\small
\begin{algorithmic}[1]
\Require Question \(q_c\), SQL \(\sigma_c\), answer \(\goldrel_s\)
\Require Base workspace \(\workspace_{\mathrm{base}}\), strategy \(h\)
\Ensure Adapted question \(q_r\), workspace \(\workspace\)
\State \(A\gets\Call{ExtractTypedAtoms}{\operatorname{AST}(\sigma_c),
    \goldrel_s,q_c}\)
\State \(\mathcal{E}_v\gets\Call{SelectAtoms}{A,h}\)
\If{\(\mathcal{E}_v=\varnothing\)}
    \State \Return \(\mathrm{reject}\)
\EndIf
\State \(S_v\gets\Call{PlanStoryboard}{\mathcal{E}_v,h}\)
\State \(\Call{ValidateAtomCoverage}{S_v,\mathcal{E}_v}\)
\State \(C_v\gets\Call{CompileComposition}{S_v}\)
\State \(a_v\gets\Call{SynthesizeNarration}{S_v}\)
\State \(v\gets\Call{RenderVideo}{C_v,a_v}\)
\State \(q_r\gets\Call{RewriteAlignedSpans}{q_c,\mathcal{E}_v,h}\)
\State \(\workspace\gets
    \Call{Integrate}{\workspace_{\mathrm{base}},v,\mathcal{E}_v,h}\)
\State \(\Call{ValidateVideoTask}{q_r,\workspace,v,\mathcal{E}_v}\)
\State \Return \(q_r,\workspace\)
\end{algorithmic}
\end{algorithm}
The normalized storyboard-planning contract appears in
Figure~\ref{fig:prompt-storyboard}.

\begin{promptblock}{Prompt for evidence-grounded storyboard planning.}
{fig:prompt-storyboard}

\begin{lstlisting}[style=appendixcode]
SYSTEM
Design a realistic data-video storyboard that communicates every supplied
evidence atom without changing its value or analytical role.

RULES
1. Every EVIDENCE_ID must appear in at least one scene and must retain its
   typed value, operator, precision, unit, and entity association.
2. Predicate abstraction must communicate both comparison direction and
   threshold. Answer evidence must remain readable but need not be adjacent.
3. Use only supplied records for data-bearing distractors. Do not invent an
   alternative rule or a second valid answer.
4. Narration should support the business scenario; do not read out all
   evidence when doing so would make visual structure unnecessary.
5. Return JSON only. Every visible constant must identify its source atom.

INPUT
TASK_STRATEGY: <PREDICATE_ABSTRACTION_OR_ANSWER_EVIDENCE>
EVIDENCE_ATOMS: <TYPED_ATOMS_WITH_STABLE_IDS>
SUPPORTING_RECORDS: <BOUNDARY_AND_DISTRACTOR_RECORDS>
VISUAL_COMPONENT_LIBRARY: <AVAILABLE_COMPONENTS>

OUTPUT
{
  "scenes": [{
    "scene_id": "...", "duration_hint": "...",
    "evidence_ids": ["..."], "component": "...",
    "data_constants": [{"atom_id": "...", "value": "..."}],
    "visible_text": ["..."], "narration": "..."
  }]
}
\end{lstlisting}
\end{promptblock}

\subsubsection{Video-Rendering Running Example}
\label{app:video-running-example}

\begin{figure*}[t]
\centering
\begin{tabular}{@{}cc@{}}
\begin{minipage}[t]{0.48\textwidth}
\centering
\includegraphics[width=\linewidth]{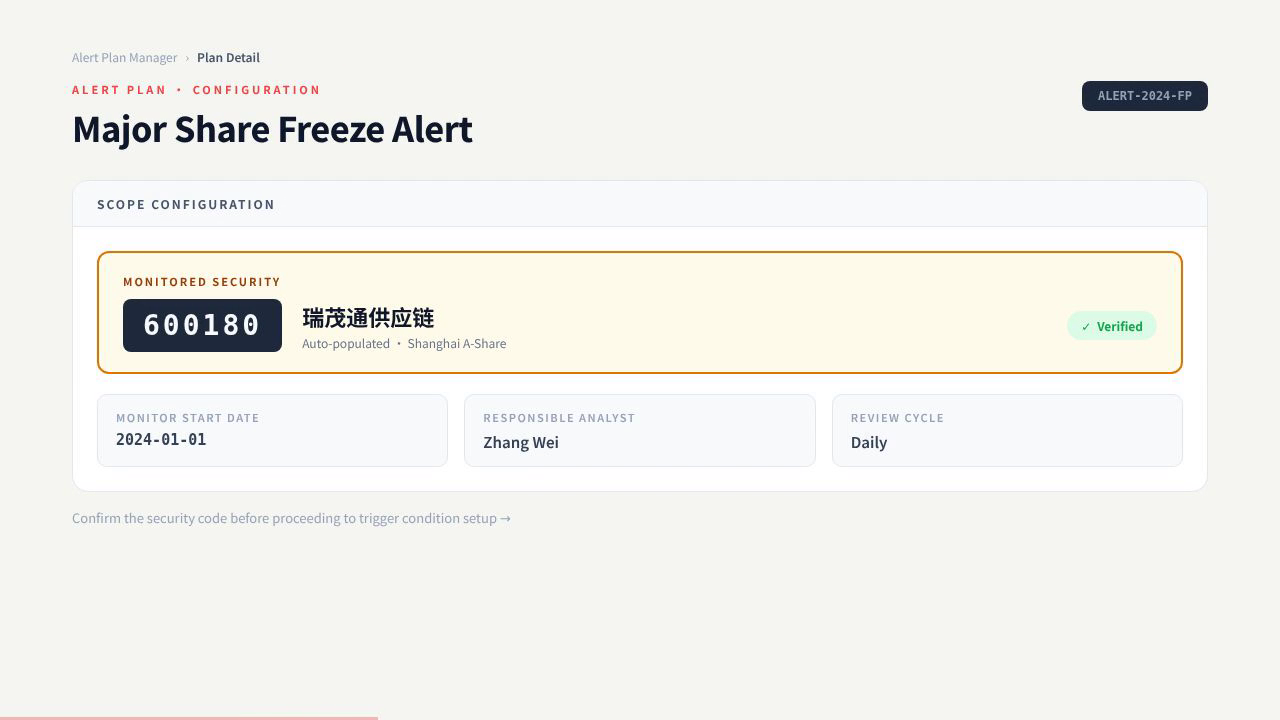}\\[-0.2em]
\small (a) Task~193: monitored-security scope
\end{minipage}
&
\begin{minipage}[t]{0.48\textwidth}
\centering
\includegraphics[width=\linewidth]{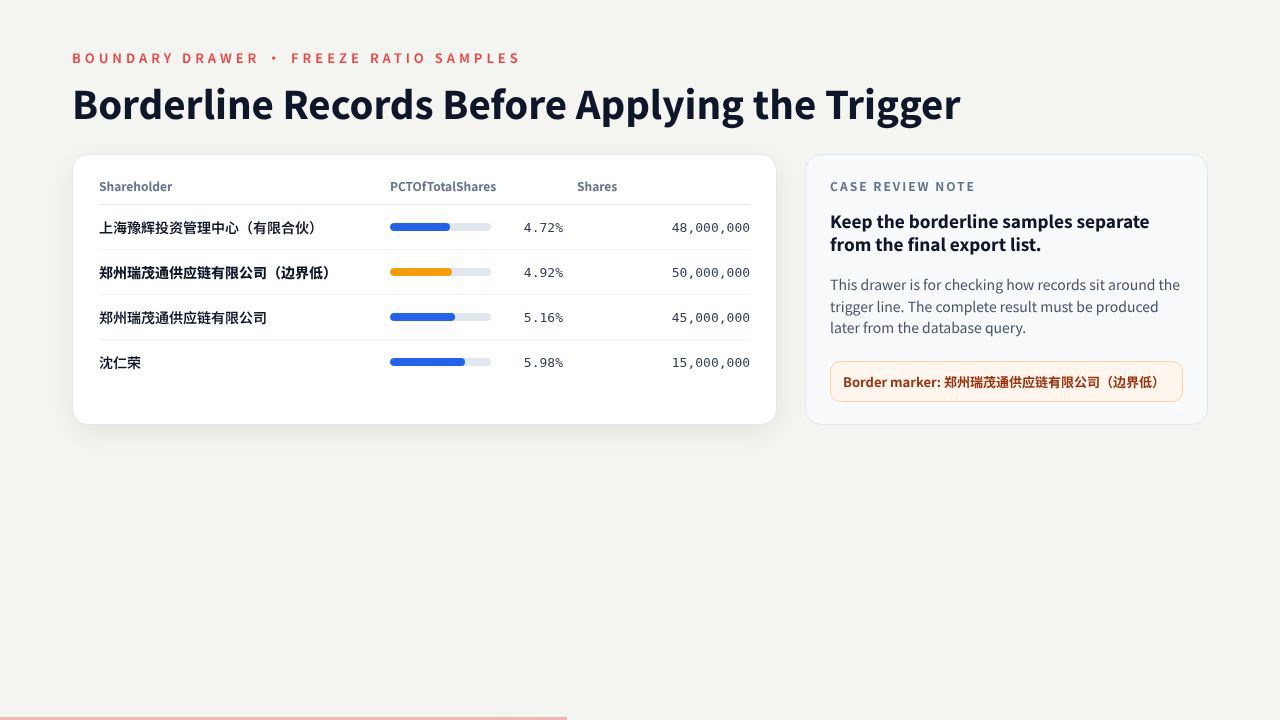}\\[-0.2em]
\small (b) Task~193: records around the decision boundary
\end{minipage}
\\[0.8em]
\begin{minipage}[t]{0.48\textwidth}
\centering
\includegraphics[width=\linewidth]{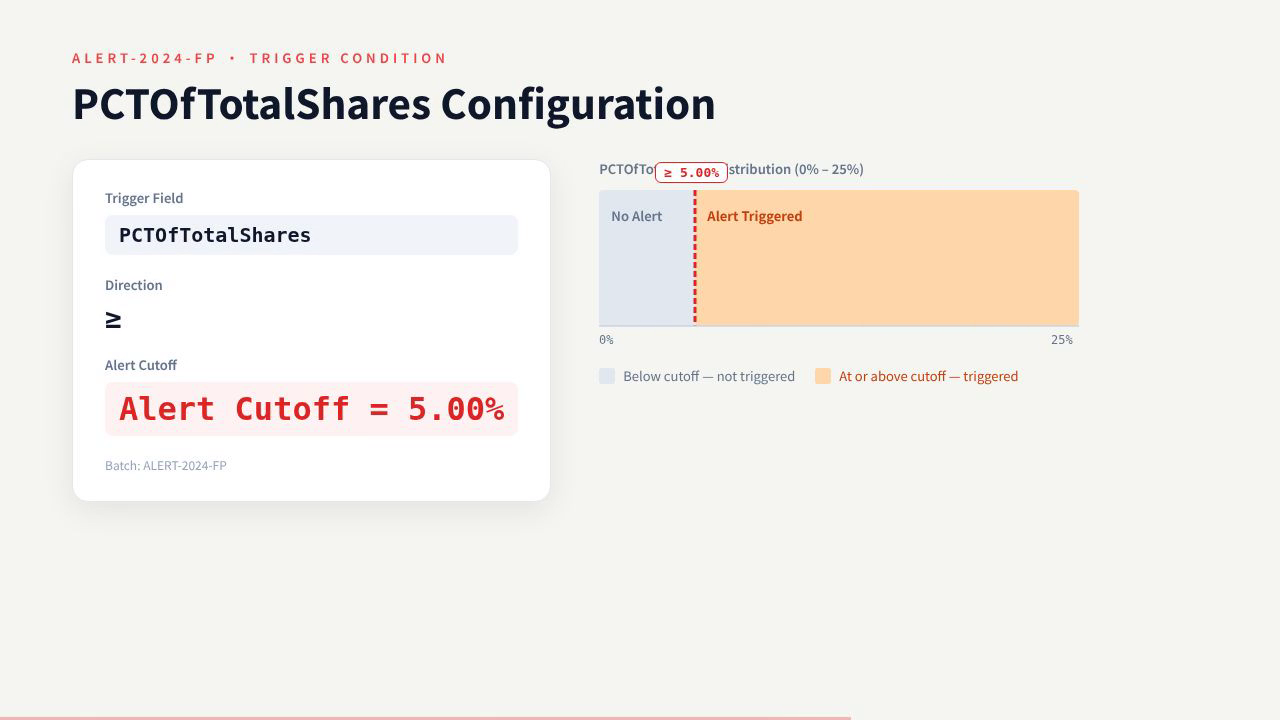}\\[-0.2em]
\small (c) Task~193: trigger field, direction, and cutoff
\end{minipage}
&
\begin{minipage}[t]{0.48\textwidth}
\centering
\includegraphics[width=\linewidth]{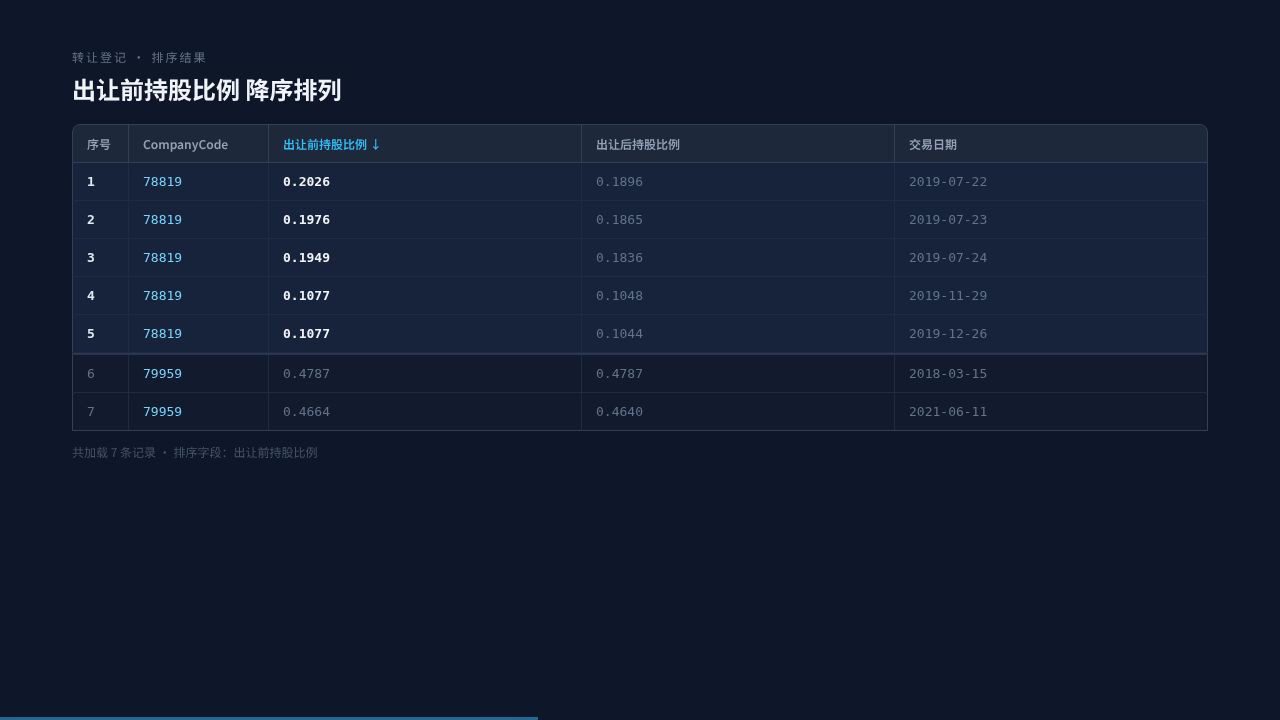}\\[-0.2em]
\small (d) Task~82: ranked answer-evidence rows
\end{minipage}
\end{tabular}
\caption{Representative frames from the released task videos. Task~193
distributes its security scope and predicate specification across separate
scenes: panel (a) identifies the monitored security, panel (b) supplies
boundary records, and panel (c) defines the trigger. Panel (d) shows the two
same-company transfer records used by the answer-evidence strategy in
Task~82.}
\label{fig:video-running-frames}
\end{figure*}

\paragraph{Predicate-abstraction example.}
Released Task~193 asks which equity-freeze records meet the ``Major Share
Freeze Alert'' configured in the video and requests shareholder, involved
shares, start date, and end date. The underlying record filter contains two
typed predicate atoms: security code \texttt{600180} and
\(\texttt{PCTOfTotalShares}\ge 0.05\). Rather than placing
these constants in the question, the video identifies the monitored security
and later displays a configuration panel whose field is
\texttt{PCTOfTotalShares}, direction is \(\ge\), and cutoff is 5.00\%.
Table~\ref{tab:video-running-example} follows these atoms into the released
task, while Figure~\ref{fig:video-running-frames}(a)--(c) shows their visual
carriers in the rendered video.

\begin{table*}[t]
\caption{Predicate-abstraction trace for released Task~193. Values in the
application row are read from the task-local SQLite artifact; the output table
omits the trigger percentage because it is not requested.}
\label{tab:video-running-example}
\small
\setlength{\tabcolsep}{5pt}
\begin{tabular}{@{}p{0.14\textwidth}p{0.82\textwidth}@{}}
\toprule
\textbf{Stage} & \textbf{Instantiated task content} \\
\midrule
Selected atoms &
\(e_1=(\texttt{SecuCode},=,\texttt{600180})\);
\(e_2=(\texttt{PCTOfTotalShares},\ge,0.05)\). The latter retains the display
form 5.00\% and the comparison direction. \\
\addlinespace
Video scenes &
An early security-monitoring scene establishes code \texttt{600180}.
Boundary examples contrast records below and above the cutoff. A later
configuration scene displays trigger field \texttt{PCTOfTotalShares},
direction \(\ge\), and ``Alert Cutoff = 5.00\%.'' \\
\addlinespace
Adapted question &
``According to the Major Share Freeze Alert configuration defined in the
video, which equity-freeze records of the monitored security meet the alert
threshold? Return the shareholder, involved shares, start date, and end
date.'' \\
\addlinespace
Application to workspace &
For security \texttt{600180}, the Shen Renrong record has involved shares
15,000,000 and \texttt{PCTOfTotalShares}=0.0598, so it passes. A neighboring
record for Zhengzhou R.M.T. Supply Chain Co. has involved shares 21,000,000
and percentage 0.024, so it fails. The agent must recover the predicate from
the video before filtering \texttt{lc\_sharefp} in SQLite. \\
\addlinespace
Reference row &
The passing example contributes
\((\text{Shen Renrong},15000000,\text{2012-04-27},\text{2012-07-24})\)
to the four-column reference table; all other passing records are returned
under the same schema. \\
\bottomrule
\end{tabular}
\end{table*}

Solving this task combines the entity scope and predicate from the video with
the freeze records in SQLite and the output projection in the question.
Neither the percentage field nor the 5.00\% cutoff is part of the requested
output.

\paragraph{Answer-evidence companion.}
Task~82 illustrates the other strategy. Its video presents two transfer rows
for company 79959 and a descending sort on pre-transfer ownership. The rows
shown are

\begin{equation}
\begin{aligned}
&(79959,\,0.4787,\,0.4787,\,\text{2018-03-15}),\\
&(79959,\,0.4664,\,0.4640,\,\text{2021-06-11}),
\end{aligned}
\end{equation}

with columns for company code, ownership before transfer, ownership after
transfer, and transaction date. The released question asks which record has
the higher pre-transfer percentage. The required output atom is therefore the
first row, while the second row supplies a same-entity comparison. The source
transfer relation is not duplicated as an agent-visible structured artifact;
the remaining workspace contains related company data and natural distractors.
This task can be solved only after reading the two percentages, associating
them with the correct rows, and applying the requested comparison.
Figure~\ref{fig:video-running-frames}(d) shows the corresponding ranked view.

\subsubsection{Artifact Validation and Failure Handling}

Validation is applied at three levels. First, renderer checks compare CSV,
JSON, and SQLite round trips with the sampled relations using
Equation~\ref{eq:artifact-roundtrip}. Second, document checks verify required
cell coverage, protected tokens, typed numeric/date/unit fidelity, null
handling, document decoding, and the absence of incompatible claims for one
record--field pair. Third, video checks verify the evidence schema, atom-to-
scene coverage, component constants, question-span replacement, compilation,
audio presence, file decoding, duration, and sampled frames.

A structural failure triggers deterministic fallback, block regeneration, or
component repair as appropriate. A semantic failure---for example, a
misstated document value, a missing comparison direction, an unreadable video
constant, or a question that still exposes relocated evidence---returns the
artifact to generation with a typed error report. Tasks that cannot satisfy
the contract within the retry budget are rejected. Accepted artifacts then
enter Human Review \& Task Repair in Section~\ref{sec:human-review}, where
reviewers inspect the complete question, workspace, and reference answer
together.

\subsection{Construction Models and Cost}
\label{app:construction-cost}

Table~\ref{tab:construction-model-cost} reports the LLM configurations and
mean API cost per task for the model-based construction components.

\begin{table}[t]
\caption{LLM configurations and API costs for benchmark construction.}
\label{tab:construction-model-cost}
\centering
\small
\setlength{\tabcolsep}{3pt}
\renewcommand{\arraystretch}{1.08}
\begin{tabular}{@{}p{0.27\columnwidth}p{0.46\columnwidth}
                    p{0.19\columnwidth}@{}}
\toprule
\textbf{Component} & \textbf{Model and role} &
\textbf{Mean USD/task} \\
\midrule
Cross-language transformation &
Qwen-Plus (generation) &
\$0.000189 \\
Document rendering &
Gemini 2.5 Pro (generation); GPT-4o (validation) &
\$0.92 \\
Video rendering &
Claude Sonnet 4.6 (generation) &
\(\sim\)\$0.58 \\
\bottomrule
\end{tabular}
\end{table}

The document cost includes generation and model-based validation. The video
cost is estimated from sampled successful video-generation records. Reported
costs cover model API usage and exclude expert review, local rendering,
storage, and other infrastructure.

\subsection{Human Review \& Task Repair Details}
\label{app:human-review-details}

Human review operates on a complete candidate task version comprising its
question, workspace, and reference result. It begins
only after the structural and artifact-level checks in
Appendix~\ref{app:artifact-rendering-details}. Those checks can establish that
files decode, database relations round-trip, and protected values are
preserved; the expert protocol determines whether the assembled task has a
clear user intent, sufficient evidence, a unique answer, and suitable
evaluation semantics.

\subsubsection{Staffing, Assignment, and Blinding}

The review pool comprises 11 domain experts with recorded domain and data-analysis
expertise. Each task is assigned to at least two reviewers. Assignments are
balanced across the active pool and can be adjusted for domain coverage before
review begins. A pending assignment may be replaced, but reviewer membership is
fixed once an independent answer has been submitted.

Reviewers, repairers, and administrators have separate roles. Reviewers solve
and verify assigned tasks; repairers prepare a new task version after a
disagreement; administrators manage assignments and record the final
accept-or-remove decision. During independent solving, a reviewer cannot access
the source SQL, reference result, discussion history, or peer submissions. Peer
judgments become visible only after that reviewer
has locked both stages of their own review.

\subsubsection{Two-Stage Review Record}

Table~\ref{tab:human-review-stages} summarizes the information boundary and
the record produced at each stage. Phase~1 requires a rectangular table
with a header; a header-only table is valid when the correct answer is
empty. Every submission must include at least one evidence reference. An
evidence reference contains an artifact path, its modality, a row, key,
section, page, or timestamp anchor, the supported fact, and its analytical
role such as filter, join, projection, aggregation, ordering, or unit
interpretation. The reviewer also records the interpreted grain, predicates,
aggregation, ordering, units, null and duplicate semantics, issue flags, and
confidence. The server validates referenced paths against the current task
version and makes the submitted result immutable.

\begin{table*}[t]
\caption{Information boundaries and required records in the expert-review
protocol. Diagnostic table comparison is shown only after the independent
answer has been locked and is not used as the official benchmark evaluator.}
\label{tab:human-review-stages}
\small
\setlength{\tabcolsep}{5pt}
\begin{tabular}{@{}p{0.16\textwidth}p{0.35\textwidth}p{0.43\textwidth}@{}}
\toprule
\textbf{Stage} & \textbf{Visible to the reviewer} & \textbf{Required record} \\
\midrule
Blind independent solving &
Final question and complete task-local workspace &
Candidate result; artifact-level evidence anchors; answer semantics; issue
flags; five-point confidence; optional rationale \\
\addlinespace
Gold verification and configuration authoring &
Locked candidate result, reference result, and diagnostic table difference &
Gold verdict; release disposition; independently authored column types,
numeric comparison rules, units, percent handling, and row-order flag; issue
flags and rationale \\
\addlinespace
Adjudication and recheck &
Cross-review summaries and discussion after all participating judgments are
locked; repaired version during independent recheck &
Evidence-grounded resolution; repair scope and reason; new-version review or
explicit removal reason \\
\bottomrule
\end{tabular}
\end{table*}

Phase~2 exposes the reference result. The reviewer
assigns a gold verdict from \textsc{Match}, \textsc{Mismatch}, and
\textsc{Unsure}, together with one disposition: \textsc{Pass}, repair the
question, workspace, gold, or configuration, repair multiple components, or
remove the task. Each reviewer independently authors a configuration covering
the semantic type of every output column, any numeric precision and percentage
convention, and whether row order is required. The configuration is derived
from the question and verified gold rather than tuned after observing model
predictions.

\subsubsection{Consensus, Adjudication, and Versioned Repair}

A task version reaches consensus only when every assigned reviewer has
completed both review stages, marked the gold as \textsc{Match}, selected
\textsc{Pass}, and submitted the same canonical evaluation configuration.
Missing verifications never count as agreement. Consequently, reviewers who
agree on all gold values still enter adjudication when their type, precision,
or ordering configurations differ.

Adjudication is evidence based rather than a majority vote. Reviewers identify
the conflicting claim and cite its location in the released workspace. The
resulting repair record states its reason and identifies which components
change: the question, workspace artifacts, reference result, evaluation
configuration, or a combination of them. Repairs follow a minimal-change
policy, retaining unaffected artifacts and semantics. Finalizing a repair
preserves the previous version, activates a new one, and reassigns the same
reviewers. The repaired version is independently rechecked before discussion
resumes, and each reviewer submits a new verification and configuration for
that version. Discussion, repair, and recheck repeat until the consensus
conditions above hold. A task that
cannot be repaired into an unambiguous and releasable instance is removed with
an explicit reason.

\subsubsection{Review Dimensions and Issue Taxonomy}

Reviewers assess structural integrity, evidence sufficiency, answer
uniqueness, gold correctness, cross-modal consistency, artifact fidelity,
video necessity, question quality, evaluation validity, and release
suitability. Table~\ref{tab:review-issue-taxonomy} groups the structured issue
flags used to make these judgments comparable across tasks. Multiple flags may
be attached to one review; free-text rationale and evidence anchors preserve
the concrete claim behind each flag.

\begin{table*}[t]
\caption{Issue taxonomy used during independent solving and gold/configuration
verification. The repair surface is selected separately, so one issue can
require changes to multiple task components.}
\label{tab:review-issue-taxonomy}
\small
\setlength{\tabcolsep}{5pt}
\begin{tabular}{@{}p{0.18\textwidth}p{0.40\textwidth}p{0.36\textwidth}@{}}
\toprule
\textbf{Category} & \textbf{Covered issues} & \textbf{Typical review question} \\
\midrule
Access and evidence &
Parsing failure, missing evidence, or mutually conflicting evidence &
Can every decisive fact be accessed and anchored in the workspace? \\
\addlinespace
Question semantics &
Ambiguous grain, filter, join, unit, time scope, ordering, null handling, or
duplicate handling &
Does the request determine one complete tabular result without relying on hidden
assumptions? \\
\addlinespace
Reference result &
Question--gold mismatch and value, shape, or order errors &
Are all and only the requested rows and columns returned with correct values? \\
\addlinespace
Artifact fidelity &
Document or video fact error, video leakage or non-necessity, translation
drift, and cross-modal inconsistency &
Do rendered artifacts faithfully carry their assigned data and reasoning role? \\
\addlinespace
Evaluation semantics &
Incorrect column type, precision, unit, percentage convention, or ordering
flag &
Does the configuration encode only distinctions required by the question? \\
\addlinespace
Release suitability &
Privacy or licensing risk, near-duplicate task, or unsolvable task &
Can the task be released and evaluated without ambiguity or prohibited data? \\
\bottomrule
\end{tabular}
\end{table*}

\subsubsection{Audit Trail and Review Interface}
\label{app:review-interface}

The review system records assignments, independent submissions,
verifications, state transitions, discussions, repair drafts and artifact
replacements, version activation, acceptance, and removal as append-only
events with actor, task version, UTC timestamp, and event payload. Base
benchmark files remain read only; repaired context files are stored as
versioned overlays. The resulting record supports agreement, issue, repair,
and post-repair acceptance statistics without reconstructing decisions from
the released files.

\begin{figure*}[t]
\centering
\begin{tabular}{@{}cc@{}}
\begin{minipage}[t]{0.48\textwidth}
\centering
\includegraphics[width=\linewidth]{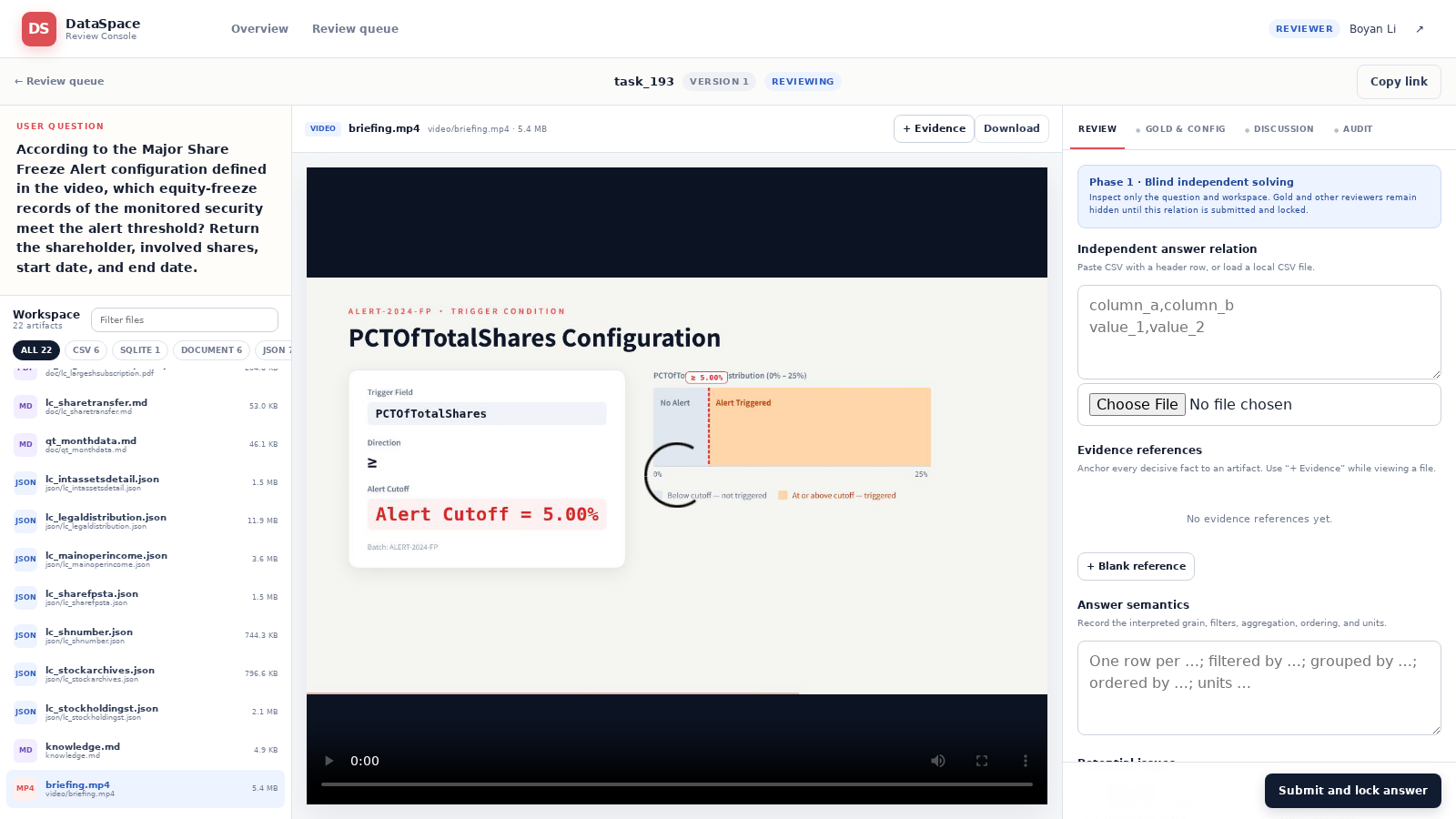}\\[-0.2em]
\small (a) Blind inspection of the heterogeneous workspace
\end{minipage}
&
\begin{minipage}[t]{0.48\textwidth}
\centering
\includegraphics[width=\linewidth]{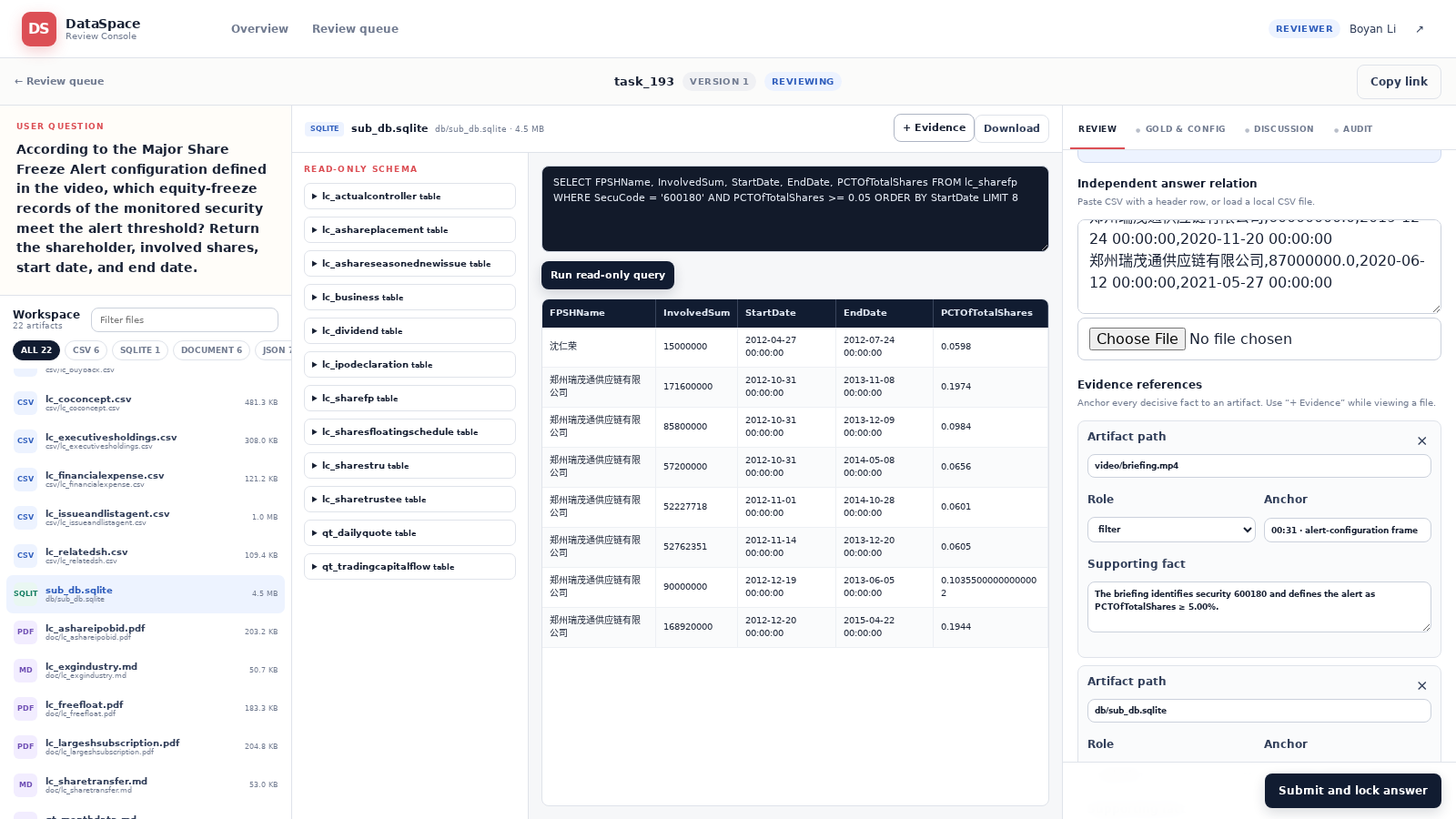}\\[-0.2em]
\small (b) Evidence-anchored independent result submission
\end{minipage}
\\[0.8em]
\begin{minipage}[t]{0.48\textwidth}
\centering
\includegraphics[width=\linewidth]{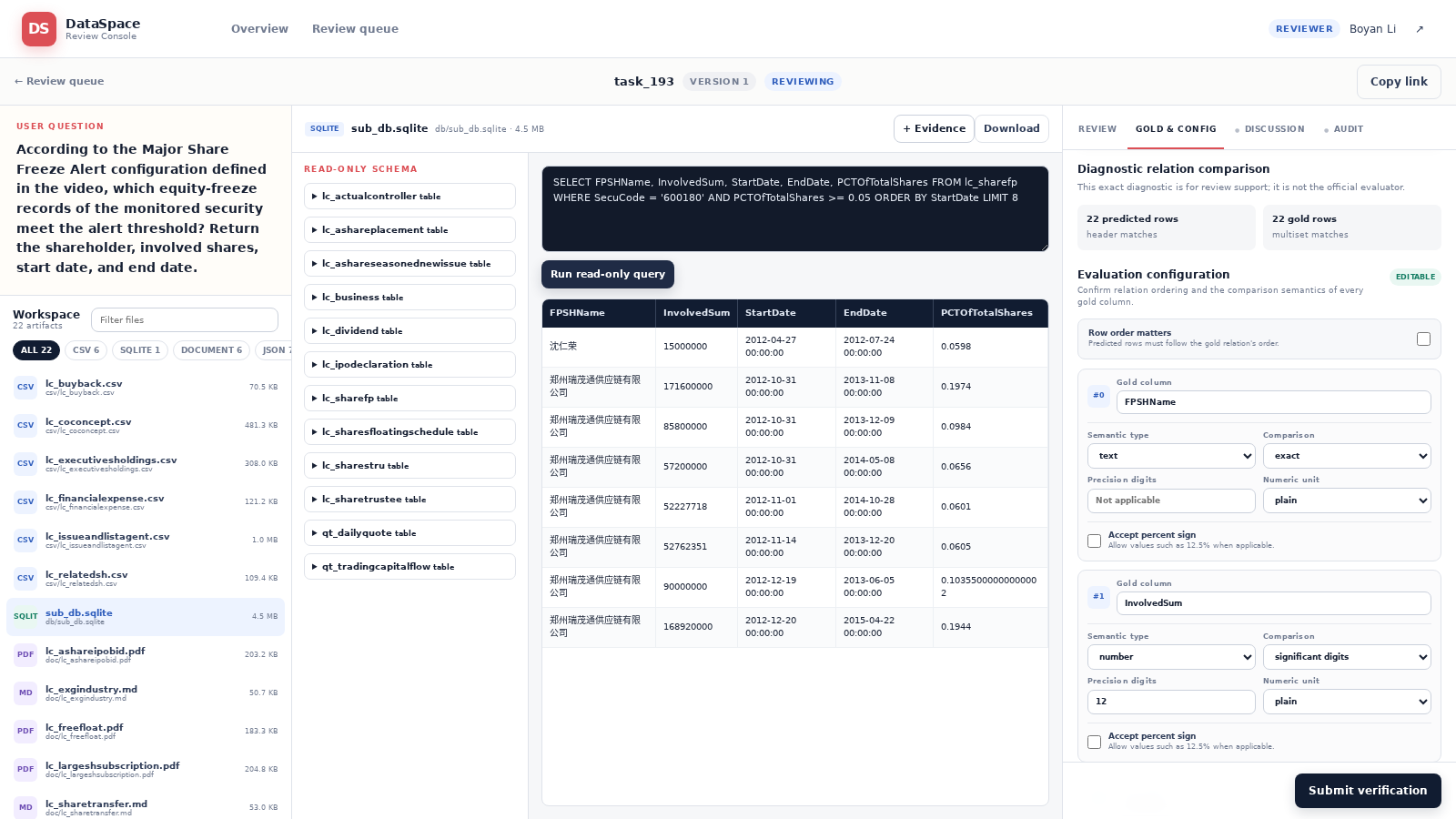}\\[-0.2em]
\small (c) Gold verification and evaluation-config authoring
\end{minipage}
&
\begin{minipage}[t]{0.48\textwidth}
\centering
\includegraphics[width=\linewidth]{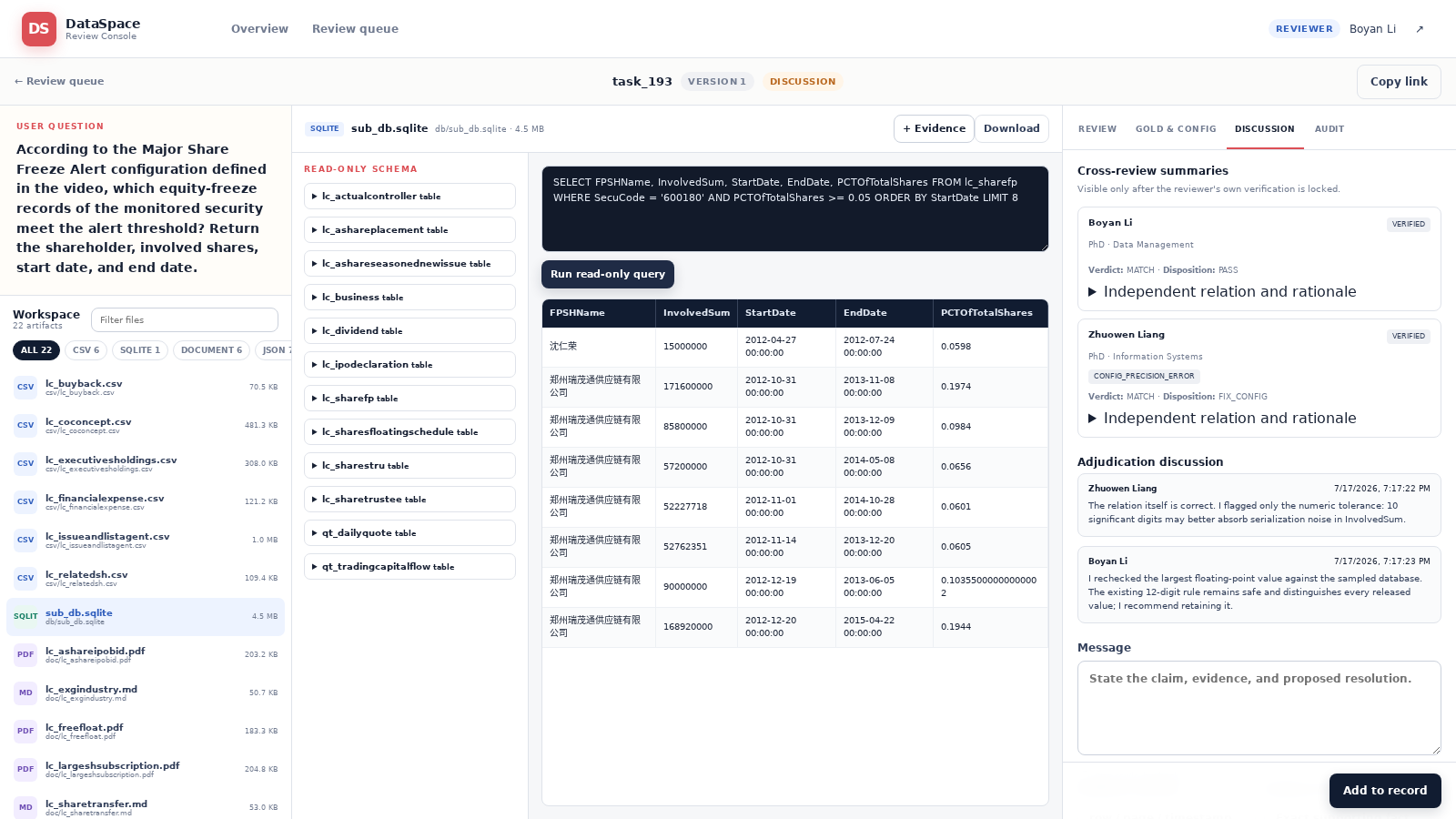}\\[-0.2em]
\small (d) Cross-review comparison and recorded adjudication
\end{minipage}
\end{tabular}
\caption{Illustrative end-to-end reviewer trace in the human-review system.
Panel (a) shows the blind phase, in which the question and complete workspace
are visible while gold and peer reviews remain locked. Panel (b) links the
candidate result to a video timestamp and filtered SQLite rows. Panel (c)
unlocks result comparison and supports independent authoring of per-column
type and precision settings after answer submission. Panel (d) exposes both
completed reviews and records a configuration disagreement for adjudication.}
\label{fig:review-system-interface}
\end{figure*}

The browser console jointly presents the question, modality-filtered workspace,
native artifact viewers, evidence form, result comparison, configuration
editor, and discussion history. Figure~\ref{fig:review-system-interface}
shows an illustrative session for released Task~193. The reviewer recovers an
alert predicate from video, applies it to SQLite, and records both evidence
anchors before the reference result is unlocked.

\subsection{Evaluation Protocol Details}
\label{app:evaluation-details}

This appendix expands the evaluation protocol in
Section~\ref{sec:evaluation}. It specifies the frozen task configuration,
canonicalization rules, joint column alignment, and tabular-result matching.

\subsubsection{Frozen Per-Task Configuration}

Each task configuration contains a schema version, the task identifier, an
ordering flag, and one entry for every reference column. Column entries are
indexed in reference order and record a human-auditable reference name and
semantic type. Numeric entries additionally specify an integer,
decimal-place, or significant-digit comparison rule; a unit convention and
permission to use a percent sign are included when relevant. The reference
name is retained for auditing but is never compared with a prediction header.

Figure~\ref{fig:evaluation-config-example} gives the released configuration
for Task~120. The task requests region, period-end date, and per-capita GDP in
the order presented by the source. Accordingly, its rows are order sensitive,
and its three columns are normalized as text, datetime, and a number rounded
to one decimal place.

\begin{figure}[t]
\centering
\begin{minipage}{\columnwidth}
\begin{lstlisting}[style=appendixcode]
{
  "schema_version": "1.0",
  "task_id": "task_120",
  "order_sensitive": true,
  "columns": [
    {
      "gold_index": 0,
      "gold_name": "Region",
      "type": "text"
    },
    {
      "gold_index": 1,
      "gold_name": "Period-end date",
      "type": "datetime"
    },
    {
      "gold_index": 2,
      "gold_name": "GDP per capita",
      "type": "number",
      "comparison": {
        "mode": "decimal_places",
        "digits": 1
      }
    }
  ]
}
\end{lstlisting}
\end{minipage}
\caption{Frozen evaluation configuration for Task~120.}
\label{fig:evaluation-config-example}
\end{figure}

\subsubsection{Canonicalization Rules}

\paragraph{Text and nulls.}
Text values are stripped of surrounding whitespace, line endings are
standardized, and Unicode is normalized to NFC; comparison otherwise remains
case sensitive. Identifiers and ordinary strings share this type because
both require exact lexical equality after normalization. An empty text cell
is canonicalized as null. For non-text columns, empty cells and the
case-insensitive tokens \texttt{null}, \texttt{none}, \texttt{nan},
\texttt{nat}, and \texttt{<na>} are treated as null. A null value matches only
another null.

\paragraph{Numbers.}
Numbers are parsed as finite base-10 decimals; ordinary decimal notation,
scientific notation, and valid comma thousands separators are accepted. The
\texttt{integer} mode requires an integral value. The
\texttt{decimal\_places} and \texttt{significant\_digits} modes round both
reference and predicted values with round-half-up at the configured
precision, after which their canonical decimal representations must be
identical.

The numeric unit is \texttt{plain}, \texttt{percentage\_points}, or
\texttt{fraction}. A percent-marked prediction is accepted only when
\texttt{allow\_percent\_sign} is enabled. Under \texttt{percentage\_points},
\texttt{3.5\%} denotes \(3.5\); under \texttt{fraction}, it denotes \(0.035\).
An unmarked value is interpreted directly in the configured unit.

\paragraph{Dates, datetimes, and Booleans.}
Dates and datetimes use ISO syntax. A date column also accepts a midnight
datetime, while a non-midnight time is invalid. Timezone-aware values are
converted to UTC before comparison. Boolean values accept
\texttt{1/true/yes/y} and their Chinese affirmative counterpart as true, and
\texttt{0/false/no/n} and the corresponding Chinese negative token as false;
alphabetic tokens are case insensitive.

\subsubsection{Joint Alignment and Tabular-Result Matching}

Prediction headers are required for valid CSV serialization but do not
participate in scoring. After checking that prediction and reference have the
same shape, the evaluator considers every one-to-one mapping from predicted
columns to reference columns. A mapping remains eligible only when all cells
in each predicted column can be canonicalized by the rule of its mapped
reference column. The mapped columns are then reassembled into rows before
comparison, so values from different prediction rows cannot be combined
independently.

For an order-sensitive task, the canonical row sequences must be identical.
Otherwise, rows are compared as multisets, retaining the multiplicity of every
duplicate row. Algorithm~\ref{alg:task-evaluation} gives the task-level
procedure used by the official evaluator.

\begin{algorithm}[t]
\caption{Task-level tabular-result evaluation}
\label{alg:task-evaluation}
\small
\begin{algorithmic}[1]
\Require Prediction \(P\), reference \(G\), configuration
    \(c=(o,\{\nu_j\}_{j=1}^{d})\)
\Ensure Binary task score
\If{\(\neg\Call{ValidCSV}{P}\)}
    \State \Return \(0\)
\EndIf
\If{\(\Call{Shape}{P}\neq\Call{Shape}{G}\)}
    \State \Return \(0\)
\EndIf
\State \(G^\star\gets\Call{Canon}{G,c}\)
\ForAll{\(\pi\in\Pi_d\)}
    \State \(P_\pi\gets\Call{ReorderColumns}{P,\pi}\)
    \If{\(\neg\Call{Compatible}{P_\pi,c}\)}
        \State \textbf{continue}
    \EndIf
    \State \(P^\star\gets\Call{Canon}{P_\pi,c}\)
    \If{\(\Call{Rows}{P^\star,o}=\Call{Rows}{G^\star,o}\)}
        \State \Return \(1\)
    \EndIf
\EndFor
\State \Return \(0\)
\end{algorithmic}
\end{algorithm}

\subsection{Experimental Configurations}
\label{app:experimental-details}

\subsubsection{Backbones}

Table~\ref{tab:backbone-configurations} records the exact endpoints used in
the controlled backbone comparison. The six models were publicly released
within the four months preceding evaluation and were accessed through Vercel
AI Gateway in July 2026. We leave reasoning effort and all unspecified
sampling parameters at the provider default; each endpoint receives the same
32,768-token output ceiling.

\begin{table*}[t]
\caption{Backbones used in the controlled \dataspaceagent comparison. Release
denotes the month of public model availability.}
\label{tab:backbone-configurations}
\small
\setlength{\tabcolsep}{6pt}
\renewcommand{\arraystretch}{1.03}
\begin{tabular}{@{}llll@{}}
\toprule
\textbf{Backbone} & \textbf{Gateway model identifier} &
\textbf{Release} & \textbf{Access} \\
\midrule
Grok 4.5~\cite{grok45} &
\texttt{xai/grok-4.5} & 2026-07 & Proprietary \\
GPT-5.6 Sol~\cite{gpt56sol} &
\texttt{openai/gpt-5.6-sol} & 2026-07 & Proprietary \\
Kimi K3~\cite{kimik3} &
\texttt{moonshotai/kimi-k3} & 2026-07 & Open-weight \\
MiMo-V2.5~\cite{mimov25} &
\texttt{xiaomi/mimo-v2.5} & 2026-04 & Open-weight \\
Claude Sonnet 5~\cite{claudesonnet5} &
\texttt{anthropic/claude-sonnet-5} & 2026-06 & Proprietary \\
MiniMax M3~\cite{minimaxm3} &
\texttt{minimax/minimax-m3} & 2026-06 & Open-weight \\
\bottomrule
\end{tabular}
\end{table*}

\subsubsection{Agent Harnesses}

The complementary harness comparison fixes the endpoint to
\texttt{xiaomi/mimo-v2.5}. Table~\ref{tab:harness-configurations} lists the
pinned implementations. All model roles exposed by a harness, including
subagents and summarization or compaction calls, are mapped to the same
backbone and model fallback is disabled.

\begin{table*}[t]
\caption{Agent harnesses compared with MiMo-V2.5 fixed as the backbone.}
\label{tab:harness-configurations}
\small
\setlength{\tabcolsep}{6pt}
\renewcommand{\arraystretch}{1.03}
\begin{tabular}{@{}lll@{}}
\toprule
\textbf{Harness} & \textbf{Version} & \textbf{Execution interface} \\
\midrule
\dataspaceagent & Ours &
Terminal-style ReAct with shell, image, and submission actions \\
Smolagents~\cite{smolagents} & 1.26.0 &
Code and tool execution in an isolated container \\
Codex~\cite{codexcli} & 0.145.0 &
Native CLI through the Responses API \\
Claude Code~\cite{claudecode} & 2.1.217 &
Native CLI with task-local configuration and state \\
Grok Build~\cite{grokbuild} & 0.2.106 &
Native CLI through the Chat Completions API \\
\bottomrule
\end{tabular}
\end{table*}

\dataspaceagent alternates model responses with tool observations until the
model submits an answer or reaches a resource limit. It exposes three actions:
\texttt{bash} executes local data-processing commands, \texttt{view\_image}
returns a selected image to the multimodal backbone, and
\texttt{submit\_answer} validates and submits a tabular result as CSV. Its prompt
specifies the task, tool interfaces, workspace location, and output contract,
without task-specific source selection, modality routing, or analytical
operators.

\subsubsection{Specialized Data-Agent Compatibility}
\label{app:specialized-agent-compatibility}

Table~\ref{tab:specialized-agent-compatibility} records their status at the
time of evaluation in July 2026. We require a system to be reproducibly
runnable over the complete task directory and to support the benchmark's
multimodal input and tabular-output contract without replacing core
components.

\begin{table*}[t]
\caption{Compatibility assessment of specialized data-agent systems. ``Public''
refers to an official implementation released by the system's authors.}
\label{tab:specialized-agent-compatibility}
\small
\setlength{\tabcolsep}{5pt}
\renewcommand{\arraystretch}{1.04}
\begin{tabular}{@{}p{0.16\textwidth}p{0.18\textwidth}p{0.60\textwidth}@{}}
\toprule
\textbf{System} & \textbf{Official implementation} &
\textbf{Reason not directly included} \\
\midrule
MLE-STAR~\cite{mlestar} & Public &
Its interface targets machine-learning competition pipelines, including web
search, model refinement, and predictive artifacts, rather than offline
analytics over heterogeneous documents and video with complete tabular
outputs. \\
Teable~\cite{teable} & Partial &
The community database platform is public, but its complete agent functions
are license-gated rather than available as a pinned, independently
reproducible implementation. It also provides no native video-analysis
interface. \\
DeepAnalyze~\cite{deepanalyze} & Public &
The released 8B model is a text-generation agent whose documented inputs cover
databases, tabular files, and text formats, but not native image or video
understanding; its primary output is an analytical report rather than a
complete tabular result. \\
TAIJI~\cite{taiji} & Unavailable &
The paper describes an MCP-based multimodal architecture and preliminary
prototype, but provides no official runnable implementation. \\
AOP~\cite{aop} & Unavailable &
No standalone implementation of the published AOP system is released; the
authors' public Unify prototype is a related but distinct system. \\
AgenticData~\cite{agenticdata} & Unavailable &
The published multi-agent planner, optimizer, executor, and memory system has
no official runnable implementation. \\
\bottomrule
\end{tabular}
\end{table*}

FDABench likewise reports reimplementing TAIJI, AOP, and AgenticData from their
papers because the original systems were not open-sourced~\cite{fdabench}.
Using such reconstructions, adding a video-capable perception model, or
rewriting a system's output path would introduce substantial choices absent
from the original systems and prevent a faithful comparison.

\subsubsection{Runtime and Reproducibility}

For the backbone comparison, each \dataspaceagent run is limited to 60 model
turns, 50 tool actions, 1,800 seconds of wall-clock time, and 180 seconds per
shell command. Its isolated runtime receives 4 CPUs, 16\,GiB of memory, and no
network access. Full-benchmark jobs use eight-way task concurrency, which
affects throughput but not per-task limits.

For the harness comparison, all systems receive a fresh session, a task-local
home directory, and Data Workbench Runtime 1.0. The runtime supplies generic
CSV, JSON, SQLite, Markdown, PDF, image/OCR, and video utilities, but no
retrieval, schema-linking, document-QA, video-QA, or Text-to-SQL solver.
Model-generated commands cannot access the network. The host-side controller
can reach only the configured inference endpoint and never exposes its
credential to generated code. Each harness receives the same 1,800-second
wall-clock limit, task input, and CSV output contract; its internal action
budget and context-management policy remain native to that harness. Prompts,
configuration files, validated predictions, and raw execution traces are
retained for release.

\subsection{Additional Statistics and Results}

\subsubsection{Backbone Efficiency}
\label{app:backbone-efficiency}

Table~\ref{tab:backbone-efficiency} reports the complete efficiency values
underlying Figure~\ref{fig:backbone-efficiency}. Cached prompt tokens are
subsets of input tokens, and reported reasoning tokens are subsets of output
tokens; neither is added again to the total. Costs are computed using the
providers' official pricing at the time of evaluation.

\begin{table*}[t]
\caption{Efficiency statistics for the controlled backbone comparison with
\dataspaceagent fixed. Token, action, and latency values are per-task
averages over all 410 tasks; parentheses give the token median and latency
90th percentile. API cost per task uses each provider's official pricing.}
\label{tab:backbone-efficiency}
\small
\setlength{\tabcolsep}{5.2pt}
\renewcommand{\arraystretch}{1.03}
\begin{tabular*}{\textwidth}{@{\extracolsep{\fill}}lrrrrr@{}}
\toprule
\textbf{Backbone} &
\textbf{Acc. (\%)} &
\textbf{Tokens (K)} &
\textbf{Cost (USD)} &
\textbf{Actions} &
\textbf{Latency (s)} \\
\midrule
Grok 4.5       & \textbf{66.34} & 301.9 (124.5) & 0.169 & 18.1 & 80.9 (185.7) \\
GPT-5.6 Sol    & 64.63          & \textbf{77.8 (50.5)} & 0.200 & \textbf{9.0} & \textbf{49.2 (90.8)} \\
Kimi K3        & 53.41          & 235.2 (100.7) & 0.235 & 19.7 & 260.1 (535.9) \\
MiMo-V2.5      & 39.27          & 237.9 (75.8) & \textbf{0.011} & 19.2 & 90.9 (219.7) \\
Claude Sonnet 5 & 32.93         & 440.4 (96.9) & 0.224 & 19.3 & 128.9 (330.2) \\
MiniMax M3     & 28.54          & 498.6 (178.6) & 0.042 & 25.0 & 104.5 (253.4) \\
\bottomrule
\end{tabular*}
\end{table*}

\noindent\textbf{Efficiency on unsuccessful tasks.}
For five of the six backbones, an incorrect task consumes between 1.2 and
3.2 times as many mean tokens as a correct task and also requires more tool
actions. GPT is the exception, with nearly identical resource use in the two
groups. Longer exploration therefore frequently reflects failure to converge
rather than additional solved tasks.

\subsubsection{Performance by Task Characteristic}
\label{app:performance-subgroups}

Table~\ref{tab:performance-subgroups} gives the sample sizes and raw
accuracies underlying Figure~\ref{fig:task-characteristic-sensitivity}. The
multimodal group pools 115 tasks requiring two modalities and 19 requiring
three.

\begin{table*}[t]
\caption{Task Accuracy (\%) by task characteristic with \dataspaceagent fixed.
Required modalities are those used by the verified solution path; workspace
quartiles are formed by total workspace bytes.}
\label{tab:performance-subgroups}
\scriptsize
\setlength{\tabcolsep}{3.8pt}
\renewcommand{\arraystretch}{0.98}
\begin{tabular*}{\textwidth}{@{\extracolsep{\fill}}llrrrrrrr@{}}
\toprule
\textbf{Characteristic} & \textbf{Group} & \textbf{\(N\)} &
\textbf{Grok} & \textbf{GPT} & \textbf{Kimi} & \textbf{MiMo} &
\textbf{Claude} & \textbf{MiniMax} \\
\midrule
\multirow{2}{*}{Language}
 & Single-language & 145 & 69.7 & 60.7 & 52.4 & 46.9 & 28.3 & 27.6 \\
 & Cross-language  & 265 & 64.5 & 66.8 & 54.0 & 35.1 & 35.5 & 29.1 \\
\midrule
\multirow{2}{*}{Required modalities}
 & Single-modal & 276 & 68.8 & 65.2 & 58.0 & 43.8 & 37.3 & 32.6 \\
 & Multimodal & 134 & 61.2 & 63.4 & 44.0 & 29.9 & 23.9 & 20.1 \\
\midrule
\multirow{2}{*}{Document evidence}
 & Absent   & 275 & 69.5 & 64.7 & 53.5 & 44.7 & 37.1 & 33.5 \\
 & Required & 135 & 60.0 & 64.4 & 53.3 & 28.1 & 24.4 & 18.5 \\
\midrule
\multirow{2}{*}{Video evidence}
 & Absent   & 313 & 66.1 & 62.0 & 52.1 & 41.5 & 33.5 & 31.3 \\
 & Required & 97  & 67.0 & 73.2 & 57.7 & 32.0 & 30.9 & 19.6 \\
\midrule
\multirow{4}{*}{Workspace size}
 & Q1 (smallest) & 103 & 78.6 & 80.6 & 69.9 & 55.3 & 52.4 & 45.6 \\
 & Q2            & 102 & 60.8 & 54.9 & 50.0 & 29.4 & 26.5 & 18.6 \\
 & Q3            & 102 & 54.9 & 62.7 & 48.0 & 34.3 & 36.3 & 29.4 \\
 & Q4 (largest)  & 103 & 70.9 & 60.2 & 45.6 & 37.9 & 16.5 & 20.4 \\
\midrule
\multirow{2}{*}{Join}
 & Absent   & 297 & 69.0 & 67.7 & 57.6 & 42.8 & 38.4 & 31.6 \\
 & Required & 113 & 59.3 & 56.6 & 42.5 & 30.1 & 18.6 & 20.4 \\
\midrule
\multirow{2}{*}{Aggregation}
 & Absent   & 264 & 64.0 & 64.0 & 52.3 & 39.8 & 35.6 & 30.7 \\
 & Required & 146 & 70.5 & 65.8 & 55.5 & 38.4 & 28.1 & 24.7 \\
\midrule
\multirow{2}{*}{Answer rows}
 & One      & 178 & 73.6 & 66.3 & 52.2 & 46.1 & 20.2 & 30.9 \\
 & Multiple & 232 & 60.8 & 63.4 & 54.3 & 34.1 & 42.7 & 26.7 \\
\midrule
\multirow{2}{*}{Answer columns}
 & One      & 197 & 65.5 & 58.4 & 45.7 & 38.1 & 15.2 & 21.3 \\
 & Multiple & 213 & 67.1 & 70.4 & 60.6 & 40.4 & 49.3 & 35.2 \\
\midrule
\multirow{2}{*}{Row order}
 & Insensitive & 318 & 63.5 & 60.4 & 49.4 & 37.4 & 25.5 & 25.5 \\
 & Sensitive   & 92  & 76.1 & 79.3 & 67.4 & 45.7 & 58.7 & 39.1 \\
\bottomrule
\end{tabular*}
\end{table*}

\subsubsection{Trace-level Root-cause Audit}
\label{app:failure-analysis}

Table~\ref{tab:root-cause-subtypes} reports the complete human-confirmed
subtype distribution used in Figure~\ref{fig:failure-analysis}.
\begin{table*}[!t]
\caption{Human-confirmed root-cause subtypes for 136 audited Grok 4.5
failures.}
\label{tab:root-cause-subtypes}
\scriptsize
\setlength{\tabcolsep}{4pt}
\renewcommand{\arraystretch}{0.94}
\begin{tabular*}{\textwidth}{@{\extracolsep{\fill}}llp{0.67\textwidth}rr@{}}
\toprule
\textbf{Stage} & \textbf{Subtype} & \textbf{Operational definition} &
\textbf{Tasks} & \textbf{Share} \\
\midrule
Q & Q1 & Target output, requested entities, or row granularity is
misunderstood. & 17 & 12.5\% \\
  & Q2 & A condition, comparison, temporal scope, or target population is
misunderstood. & 13 & 9.6\% \\
  & Q3 & An ordering or answer constraint is misunderstood. & 1 & 0.7\% \\
\addlinespace[1pt]
D & D2 & The wrong artifact or source is selected as authoritative evidence.
& 3 & 2.2\% \\
\addlinespace[1pt]
E & E1 & A read, retrieval window, page, frame, or query omits required
records. & 5 & 3.7\% \\
  & E2 & The representation of the correct artifact is parsed incorrectly.
& 3 & 2.2\% \\
  & E3 & Document, image, or video evidence is transcribed incorrectly.
& 1 & 0.7\% \\
\addlinespace[1pt]
G & G1 & A field or schema element is assigned the wrong meaning.
& 5 & 3.7\% \\
  & G2 & Entities, identifiers, join keys, or records are aligned incorrectly.
& 1 & 0.7\% \\
  & G3 & A correctly read value is normalized with the wrong unit, date,
language, or scale. & 2 & 1.5\% \\
  & G4 & Conflicting sources or versions are reconciled incorrectly.
& 4 & 2.9\% \\
\addlinespace[1pt]
C & C1 & Filtering, Boolean, or NULL logic is applied incorrectly.
& 2 & 1.5\% \\
  & C2 & A join, set operation, or deduplication step is incorrect.
& 1 & 0.7\% \\
  & C3 & Aggregation, grouping, windowing, or ranking is incorrect.
& 1 & 0.7\% \\
  & C6 & A corrected intermediate result is not propagated through a
multi-step computation. & 1 & 0.7\% \\
\addlinespace[1pt]
M & M1 & A correct internal result is submitted with extra or missing
columns. & 60 & 44.1\% \\
  & M2 & A correct internal result is submitted with extra or missing rows.
& 1 & 0.7\% \\
  & M3 & Type, numeric precision, date precision, or value formatting is
altered during output. & 9 & 6.6\% \\
  & M4 & The correct rows are serialized in an incorrect required order.
& 1 & 0.7\% \\
\addlinespace[1pt]
T & T1 & Unproductive iteration exhausts the action budget without an earlier
persistent analytical error. & 4 & 2.9\% \\
  & T3 & The agent terminates or fails to submit after obtaining the answer.
& 1 & 0.7\% \\
\midrule
\multicolumn{3}{r}{\textbf{Total}} & \textbf{136} & \textbf{100.0\%} \\
\bottomrule
\end{tabular*}
\end{table*}

The audit unit is one failed Grok 4.5 task. We assign exactly one primary
cause: the earliest observable divergence that conflicts with the verified
solution, remains uncorrected, and determines the submitted result or
prevents submission. Exploratory errors that the agent later corrects are not
primary causes, and evaluator outcomes such as a column-count mismatch or
missing prediction are retained only as symptoms.

\paragraph{Audit procedure.}
For each of the 136 audited failures, GPT-5.6 Sol receives the observable
tool/action trace, workspace, submitted prediction, reference result, evaluation
configuration, and verified solution annotation. It proposes a primary
category and subtype, cites the trace event at which the persistent divergence
first appears, and gives a counterfactual correction. A human researcher
checks every cited event and the relevant workspace evidence, then confirms
or revises the proposal. The audit does not use or claim access to hidden
chain-of-thought content.

\paragraph{Stage boundaries.}
\emph{Task intent} (Q) covers an incorrect formulation of the requested
output, conditions, scope, or ordering constraint. \emph{Discovery} (D)
covers failure to locate the required evidence or selection of the wrong
source. \emph{Extraction} (E) applies when the correct artifact is accessed
but its raw values are not recovered accurately; \emph{grounding} (G) applies
when those values are read but assigned the wrong field, entity, unit, or
source interpretation. \emph{Computation} (C) requires correct evidence and
semantics followed by an incorrect relational or numerical operation.
\emph{Materialization} (M) is used only when the trace already contains the
correct target schema and internal table, or values directly projectable
to it. \emph{Termination} (T) is used only when no earlier persistent
Q/D/E/G/C error explains the failure.

\section{Competition Deployment and Artifact Availability}
\label{sec:artifact-availability}

\noindent\textbf{Competition deployment.}
\dataspace served as the official evaluation benchmark for the KDD Cup 2026
Data Agents for Complex Data Analysis competition~\cite{kddcup2026dataagents}.
The competition evaluated containerized agent systems through hidden A- and
B-board evaluation under challenge-specific runtime, submission, and
leaderboard rules. The release described in this paper instead uses the
finalized semantics-aware evaluation protocol in
Section~\ref{sec:evaluation}.

\noindent\textbf{Artifact availability.}
All 410 task inputs are publicly available at
\url{https://huggingface.co/datasets/HKUSTDial/DataSpace}.
The release includes reference answers and evaluation configurations for 60
representative tasks, enabling local end-to-end evaluation; the remaining 350
references are withheld for official full-benchmark evaluation. The official
evaluator, baseline implementations, experiment configurations, and
documentation are available at
\url{https://github.com/HKUSTDial/DataSpace}. Both repositories are released
under the MIT License.

\end{document}